\documentclass[10pt,journal,compsoc]{IEEEtran}

\ifCLASSOPTIONcompsoc
  \usepackage[nocompress]{cite}
\else
  \usepackage{cite}
\fi

\usepackage{url}
\usepackage{color}
\usepackage{stfloats}
\usepackage{graphicx}
\usepackage{float}

\usepackage{amsmath,amsfonts}
\usepackage{algorithmic}
\usepackage{algorithm}
\usepackage{array}
\usepackage[caption=false,font=normalsize,labelfont=sf,textfont=sf]{subfig}
\usepackage{textcomp}
\usepackage{stfloats}
\usepackage{url}
\usepackage{verbatim}
\usepackage{graphicx}
\usepackage{cite}
\usepackage{float}
\usepackage{CJKutf8}
\usepackage{amssymb}
\usepackage{times}
\usepackage{soul}
\usepackage{ragged2e}
\usepackage{multirow}
\usepackage{booktabs}
\usepackage{lineno}
\usepackage{pifont}
\usepackage{color}
\usepackage{float}
\usepackage{ragged2e}
\usepackage{makecell}
\usepackage{hyperref}
\usepackage{framed}

\begin{document}
\title{Is Multi-Modal Necessarily Better? Robustness Evaluation of Multi-modal Fake News Detection}

\author{Jinyin Chen,~\IEEEmembership{}
        Chengyu Jia,~\IEEEmembership{}
        Haibin Zheng,~\IEEEmembership{}
        Ruoxi Chen~\IEEEmembership{}
        and Chenbo Fu~\IEEEmembership{}% <-this % stops a space
% \IEEEcompsocitemizethanks{\IEEEcompsocthanksitem M. Shell was with the Department
% of Electrical and Computer Engineering, Georgia Institute of Technology, Atlanta,
% GA, 30332.\protect\\
\IEEEcompsocitemizethanks{\IEEEcompsocthanksitem This research was supported by the National Natural Science Foundation of China (No. 62072406), the National Key Laboratory of Science and Technology on Information System Security (No. 61421110502), the Key R\&D Projects in Zhejiang Province (No. 2021C01117), the 2020 Industrial Internet Innovation Development Project (No.TC200H01V) and ``Ten Thousand Talents Program" in Zhejiang Province (No. 2020R52011).\protect\\}
% % note need leading \protect in front of \\ to get a newline within \thanks as
% % \\ is fragile and will error, could use \hfil\break instead.
% E-mail: see http://www.michaelshell.org/contact.html
% \IEEEcompsocthanksitem J. Doe and J. Doe are with Anonymous University.}% <-this % stops an unwanted space
\thanks{J. Chen is with the Institute of Cyberspace Security, and the College of Information Engineering, Zhejiang University of Technology, Hangzhou, 310023, China. (e-mail: chenjinyin@zjut.edu.cn)}
\thanks{C. Jia is with Zhejiang University of Technology, Hangzhou, 310023, China. (e-mail: kenan976431@163.com)}
\thanks{H. Zheng is with Zhejiang University of Technology, Hangzhou, 310023, China. (e-mail: haibinzheng320@gmail.com)}
\thanks{R. Chen is with Zhejiang University of Technology, Hangzhou, 310023, China. (e-mail: 2112003149@zjut.edu.cn)}
\thanks{Corresponding author: C. Fu is with the Institute of Cyberspace Security, and the College of Information Engineering, Zhejiang University of Technology, Hangzhou, 310023, China. (e-mail: cbfu@zjut.edu.cn)}}

\markboth{IEEE TRANSACTIONS ON NETWORK SCIENCE AND ENGINEERING}%
{Shell \MakeLowercase{\textit{et al.}}: Bare Demo of IEEEtran.cls for Computer Society Journals}

\IEEEtitleabstractindextext{%
\begin{abstract}
The proliferation of fake news and its serious negative social influence push fake news detection methods to become necessary tools for web managers. Meanwhile, the multi-media nature of social media makes multi-modal fake news detection popular for its ability to capture more modal features than uni-modal detection methods. However, current literature on multi-modal detection is more likely to pursue the detection accuracy but ignore the robustness of the detector. To address this problem, we propose a comprehensive robustness evaluation of multi-modal fake news detectors. In this work, we simulate the attack methods of malicious users and developers, i.e., posting fake news and injecting backdoors. Specifically, we evaluate multi-modal detectors with five adversarial and two backdoor attack methods. Experiment results imply that: (1) The detection performance of the state-of-the-art detectors degrades significantly under adversarial attacks, even worse than general detectors; (2) Most multi-modal detectors are more vulnerable when subjected to attacks on visual modality than textual modality; (3) Popular events' images will cause significant degradation to the detectors when they are subjected to backdoor attacks; (4) The performance of these detectors under multi-modal attacks is worse than under uni-modal attacks; (5) Defensive methods will improve the robustness of the multi-modal detectors.
%The data and code are available at \url{https://github.com/kenan976431/Robustness_Multi-modal_Detector}.
\end{abstract}

}

\maketitle

\IEEEdisplaynontitleabstractindextext

% \fi
\IEEEpeerreviewmaketitle

\IEEEraisesectionheading{\section{Introduction}\label{sec:introduction}}

\IEEEPARstart{T}{he} popularity of social media has deeply affected the way people consume information. However, the accompanying risks, e.g., spreading fake news more easily continue increasing. The deep entanglement online and offline makes fake news as dangerous as a fast-inflating bubble. For example, during the 2016 U.S. presidential election, fake news related to the two candidates was shared more than 37 million times on Facebook~\cite{farajtabar2017fake}. Moreover, during the outbreak of the COVID-19, lots of fake news about this pandemic on social media have harmed people's health-protective behaviors~\cite{van2020inoculating}. 
%Recently, photos of past wars spread in the news of the Ukrainian war, to exaggerate the situation of the war as a political tool.

In the aspect of context style, social media attracts users not only with traditional text but also images and short videos, which provides a better reading experience and credibility. Unfortunately, malicious users can still abuse this multi-media information. Unlike text-only information, malicious users on social media can manipulate information in more imperceptible ways, such as fake photos, unrelated images, caricatures, etc.
Moreover, fake news with multi-modal information usually has a faster spreading speed and negative effect~\cite{wang2018eann}. Consequently, text-based detection methods are challenged by multi-modal information, leading to unsatisfying detection accuracy~\cite{shu2017fake}. Under such a circumstance, fake news detection on social media (mostly multi-modal information) has recently become an emerging research topic. On the one hand, researchers have conducted fake news detection methods based on multi-media content~\cite{jin2016novel} which have achieved better performance. On the other hand, assisted the manual fact-checking methods, fact-checking websites are emerged to help people distinguish fake news, such as Snopes, FactCheck, PolitiFact and Full Fact. However, to achieve high accuracy, these systems usually have a high cost of manual effort, e.g., manual annotation or fact-checking.

\begin{figure*}[!t]
\centering
\subfloat[]{\includegraphics[width=3.5in]{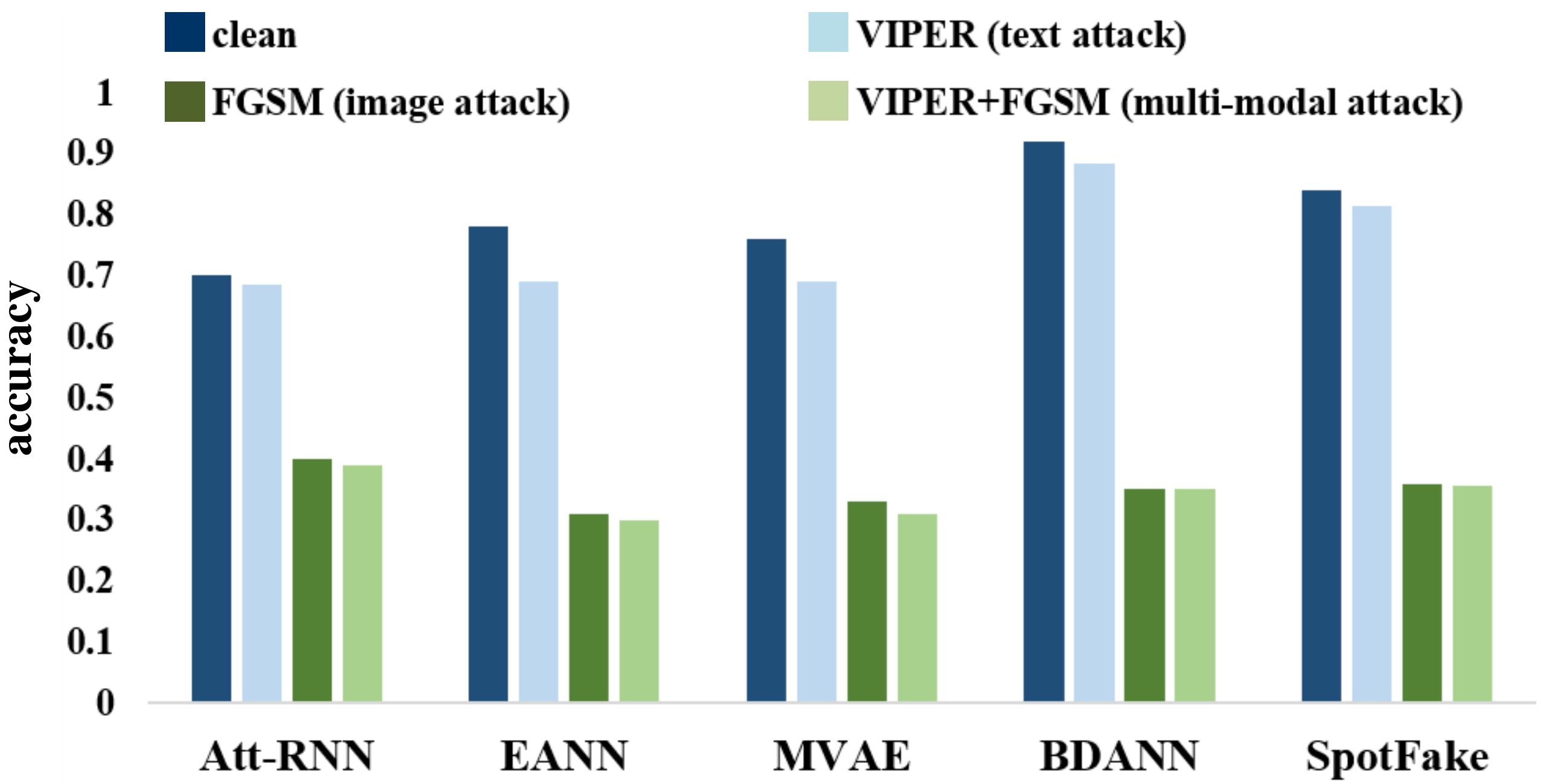}%
\label{multi1}}
\hfil
\subfloat[]{\includegraphics[width=2.5in]{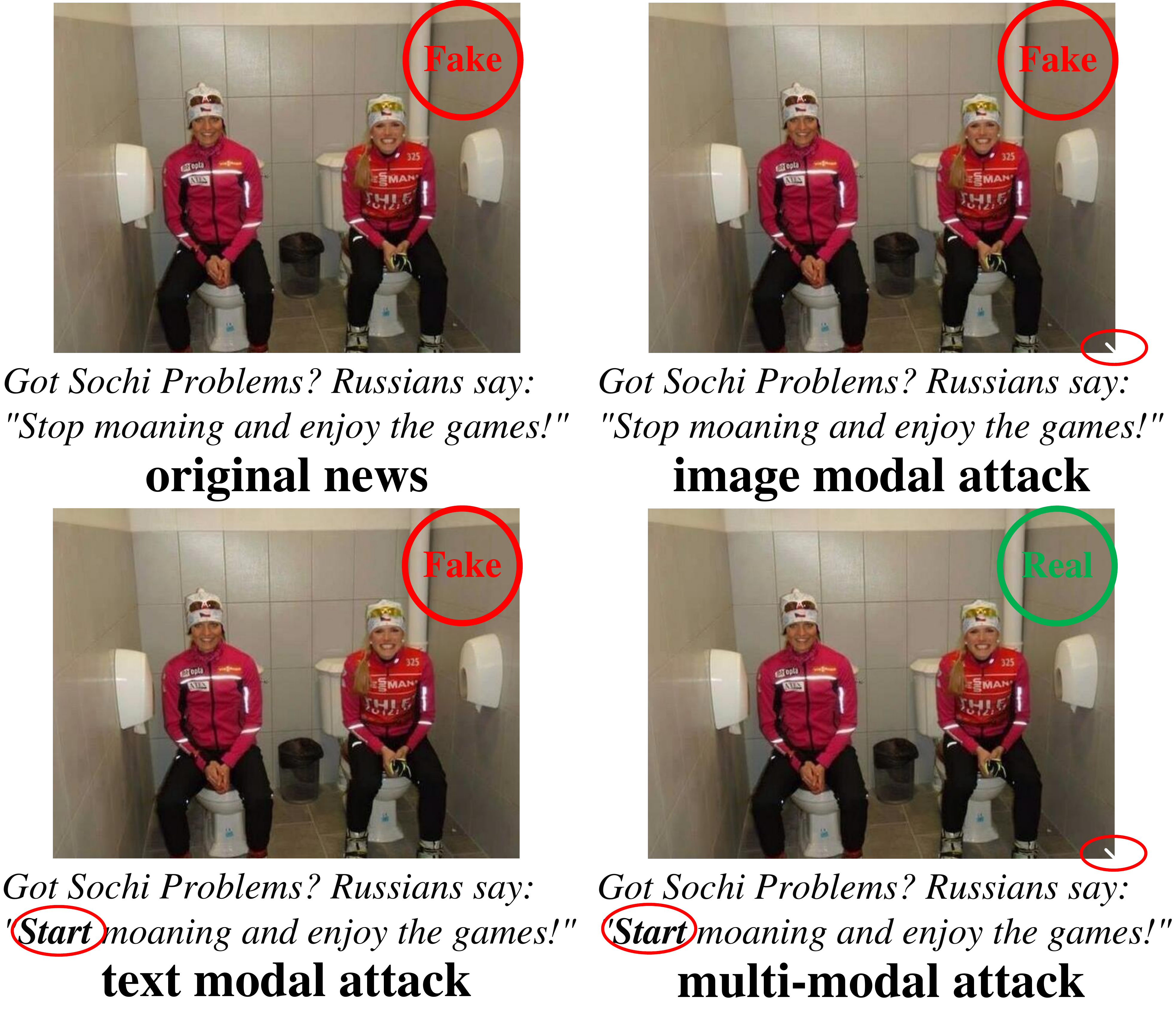}%
\label{multi2}}
\caption{The specialties between multi-modal and uni-modal attacks. (a) Detectors' performance under multi-modal and uni-modal attacks. Using the Twitter~\cite{boididou2015verifying} as the dataset, 
the perturbation for both VIPER\cite{2019Text} and FGSM~\cite{goodfellow2015explaining} is set to 0.1. Specifically, `clean' denotes the detection performance when dealing with original datasets before attacks, `text attack' and `image attack' represent the detection performance under adversarial attack on text and image alone, respectively, and `multi-modal attack' is the attack on both two modals. (b) Multi-modal attacks make detectors identify errors. The original fake news in the upper left corner can be detected normally. The upper right corner adds a patch on the fake news image, and the lower left corner replaces a word in the fake news text. Uni-modal attack on the image cannot help the fake news bypass detector's detection. The lower right corner is the fake news after a multi-modal attack that includes both, escaping the detection.}
\label{fig:multi}
\end{figure*}

The rapid development of multi-modal detector methods exhibits the dynamic game process between website managers and malicious users (developers). In order to achieve specific political or economic benefits, the malicious users or developers will do their best to deceive the detectors. For example, substituting subtle synonyms or similar words can make the text misclassified in natural language processing (NLP) tasks~\cite{ren2019generating}. 
According to the stage that the attack is conducted, two types of attacks have been introduced, including \emph{adversarial attack}, i.e., imperceptible perturbation added to the data in the testing process, to fool the model to output the wrong result, and \emph{backdoor attack}, i.e., specifically designed trigger added to some of the data in the training process, to make the model output the targeted result when fed by some triggered examples.
It has been widely proved that the imperceptible perturbation in images can make the classifier fail in computer vision tasks~\cite{goodfellow2015explaining}. Besides, malicious developers may introduce backdoor attacks in outsourced training scenarios~\cite{gu2019badnets, dai2019backdoor}.
Consequently, due to the inherent vulnerability of deep models (the nature of being vulnerable to adversarial attacks and backdoor attacks)~\cite{carlini2017towards,gu2019badnets}, these multi-modal detectors based on deep model feature extractors are also vulnerable to interference from both adversarial attacks and backdoor attacks. Therefore, the robustness of these deep neural models becomes important for it represents the ability to maintain the performance of the main task under both clean and attacked scenarios. 
The issue of adversarial attack on text-based fake news detectors~\cite{ali2021all} has been explored, but it does not consider robustness in multi-modal detectors and other scenarios.

To better illustrate the robustness of the current dominant multi-modal fake news detectors (attention-based recurrent neural network (Att-RNN)~\cite{liu2016attention}, event adversarial neural networks (EANN)~\cite{wang2018eann}, multi-modal variational auto-encoder (MVAE)~\cite{khattar2019mvae}, BERT-based domain adaptation neural network (BDANN)~\cite{zhang2020bdann} and SpotFake~\cite{singhal2019spotfake}), we evaluate their detection accuracy before and after being attacked by three adversarial attacks, i.e., visual perturber (VIPER)~\cite{2019Text}, image-based adversarial attack named fast gradient sign method (FGSM)~\cite{goodfellow2015explaining} and multi-modal attack use both attack methods. The fake news detection accuracy comparison results of these detectors before and after attacks are shown in Fig.~\ref{fig:multi} (a). 
%The ordinate represents the original detection accuracy of the five detectors and the detection accuracy under different modal adversarial attacks. 
It can be easily observed that all five detectors are performing well for clean news, i.e., more than 70\%. However, when under attacks, all of them sharply decreased near to 40\%, which is solid evidence to prove that malicious attackers may attack both modalities simultaneously if they wish to keep their fake messages evading detection by these detectors.
Fig.~\ref{fig:multi} (b) is an example of the fake news carefully crafted to bypass the detector of MVAE~\cite{khattar2019mvae}. The fake news cannot deceive the detector only with a uni-modal attack, but it will be falsely detected when subjected to a multi-modal attack.

Another robustness issue of the deep learning model has also captured our attention, 
named \emph{biased deep learning}. 
In this work, bias in multi-modal detection refers to that the detector pays more attention to one modality (e.g., image) than another (e.g., text). 
The detector with strong bias is more vulnerable, which needs only half or even lower perturbation cost to be attacked. The barrel effect means that the robustness of the multi-modal detector depends on the robustness of the short plate modality.

Consequently, it is necessary to comprehensively study the robustness of multi-modal detectors before practical deployment in the real world. In this work, we conduct a comprehensive robustness evaluation of the multi-modal fake news detectors to address these problems. Specifically, we evaluate fake news detection models, focusing on four research questions (RQs).
\begin{itemize}
\item \textbf{RQ1}: How robust are the well-performing multi-modal detectors under adversarial attacks (attacks by malicious users)?
\item \textbf{RQ2}: How do backdoor attacks (attacks by malicious developers) affect the robustness of multi-modal detectors?
\item \textbf{RQ3}: Are the multi-modal detectors biased (which modality affects the detector more)?
\item \textbf{RQ4}: Can the robustness of these multi-modal detectors be improved (defend against malicious attacks and deal with special scenarios)?
\end{itemize}

To answer these research questions, we select five multi-modal fake news detection methods with dominant performances, i.e., Att-RNN, EANN, MVAE, BDANN, and SpotFake. First, we record their detection accuracy and try to explain their behaviors under both clean and attack conditions through several level interpretation tools, i.e., 
% in the perceptive of salient feature captured by the detector through Grad-CAM visualization~\cite{selvaraju2017grad},
% neural level activation features from activated pathways~\cite{jin2021catchbackdoor},
latent textual feature representations~\cite{van2008visualizing} learned by these detectors. Furthermore, we compare their detection performance changes before and after the adversarial attack (test phase) to answer RQ1.
Second, we compare the detection performance of clean detectors and backdoored detectors to answer RQ2. 
Then, for RQ3, we attack textual, visual and multi-modal features extractor, respectively, as well as the detector's detection experiments in the case of image data style transfer. In the condition of visual and textual data mismatch. 
At last, for RQ4, based on the conclusion of RQ1 and RQ2, we utilize two common methods of defense to testify the possibility of robustness improvement for these detectors.

The main contributions of our work are summarized as follows: 

\begin{itemize}
\item To the best of our knowledge, this is the first work to perform a comprehensive robustness evaluation on multi-modal fake news detectors (i.e., adversarial attack, backdoor attack, and biased evaluation).
\item We analyze the robustness of multi-modal fake news detectors under various attacks to simulate malicious users and developers, and conclude novel insights from extensive experiments.
\item We propose defensive methods to improve the robustness of multi-modal fake news detectors, i.e., image resizing, adversarial training, and activation clustering based defenses.
\end{itemize}

The remaining part of this paper is organized as follows:
Related works are introduced in Section~\ref{related}, while preliminaries and critical method are detailed in Section~\ref{pre} and~\ref{method}. Experiments and analysis are shown in Section~\ref{exp}. In Section~\ref{discussion}, we discuss robustness in special scenarios. Finally, we conclude our work and discuss limitations in Section~\ref{conclusion}.

\section{Related Work\label{related}}
This section briefly reviews the related works of multi-modal fake news detectors, adversarial attacks, backdoor attacks, and modality bias in deep learning.

\subsection{Multi-modal Fake News Detectors}
Traditional fake news detection models mostly rely on texts, which utilize statistical and semantic features from the text content~\cite{kumar2016disinformation,song2019ced}, or statistical analysis of communication based on social networks~\cite{shrivastava2020defensive}. In order to extract more effective features, recent studies focus on multi-modal contents. 
For example, Jin et al.~\cite{jin2017multimodal} used deep neural networks to fuse multi-modal content on social networks. They proposed that the Att-RNN method used the attention mechanism to fuse multi-modal contents. Wang et al.~\cite{wang2018eann} built an end-to-end model for fake news detection and event discriminator, namely EANN. It could remove the features of specific events that couldn't migrate, and retained the shared features between events to detect fake news. Inspired by the EANN model, Khattar et al.~\cite{khattar2019mvae} built a similar architecture named MVAE. It utilized a bi-modal variational autoencoder and binary classifier for fake news detection. Similarly, inspired by the event classifier~\cite{wang2018eann} and the domain adaptive~\cite{khattar2019mvae}, Zhang et al.~\cite{zhang2020bdann} introduced a domain classifier to remove the dependency of specific events from the features extracted by the multi-modal features extractor and proposed the BDANN framework. It used the bidirectional encoder representations for transformers (BERT) and visual geometry group (VGG19) models to extract textual and visual features, respectively. The SpotFake~\cite{singhal2019spotfake} framework also used BERT and VGG19, which was proposed to solve the problem that the results of fake news detection rely heavily on subtasks, and didn't consider any other subtasks to detect fake news effectively.

In summary, the existing works of multi-modal fake news detection mainly focused on detection performance but ignored the robustness of these methods under adversarial circumstances.

\subsection{Adversarial Attacks}
In this section, we briefly introduce the works relate to adversarial attacks on images and texts. The adversarial attack is designed to deceive the artificial intelligence systems, and to simulate the malicious users' attack action by adding adversarial pixels to images or replacing words and characters in the text.

\subsubsection{Adversarial Attacks on Images}
Adversarial attacks originated in the field of computer vision. The large BFGS (L-BFGS) method proposed by Szegdy et al.~\cite{szegedy2016rethinking} solved the optimization problem of misleading the model for the adversarial examples of the image classification task. Although L-BFGS was effective, the computational cost was high, which inspired Goodfellow et al.~\cite{goodfellow2015explaining} to propose a simpler solution, namely FGSM. This method setted the perturbation as the product of the gradient sign and the step size, which increased the loss of the model. Different from the gradient attack used by FGSM, the Jacobian-based saliency map attack (JSMA) proposed by Papernot et al.~\cite{papernot2016limitations} used the Jacobian matrix of the neural model to evaluate the output sensitivity of the neural model to each input component, and gave greater control to the adversarial examples under the given perturbation. DeepFool~\cite{moosavi2016deepfool} was an iterative L2 regularization algorithm. Projected gradient descent (PGD) reduced the attack and defense into the min-max optimization framework. It assumed that the neural network is linear, so the hyperplane could be used to distinguish classification.

\subsubsection{Adversarial Attacks on Texts}
Due to the inherent differences between visual and textual data, countermeasures for images can't be directly applied to text data. 
Ebrahimi et al.~\cite{ebrahimi2018hotflip} proposed a character-level attack method HotFlip, which used the directional derivative represented by one-hot input to estimate which character to replace, and combined beam search to find the right combination of character changes. Jia and Liang~\cite{jia2017adversarial} generated adversarial examples by adding some meaningless sentences at the end of the paragraph. Gao et al.~\cite{morris2020textattack} proposed DeepWordBug generate adversarial examples against recurrent neural network (RNN) models, which used a scoring function to calculate the importance of words in a sequence under a black-box scenario and character-level modifications to make spelling mistakes. Since spelling errors were easy to detect and correct, Jin et al.~\cite{jin2019bert} proposed a black-box attack method TextFooler, which performed synonym substitution for important words and checked the semantic similarity of sentences to fool the system. Currently, most studies of text adversarial attacks are based on English data, which is not suitable for Chinese data. Wang et al.~\cite{wang2020towards} proposed a Chinese adversarial example generation method. This method replaced homophones in the Chinese input text in a black-box scenario, effectively changing the tendency of long-short term memory (LSTM) and convolutional neural network (CNN) models to classify the modified examples.

% Furthermore, although these methods (adversarial attacks on images or text) seem complex, their application is simple because these models point out vulnerable words or pixels. We want to explore whether the performance of multi-modal detectors based on deep learning feature extractors is affected by such attacks.

\subsection{Backdoor Attacks}
Similar to the adversarial attack, the backdoor attack simulates the malicious developers' attack action by adding watermarks and pixel blocks to images or adding fixed strings to text.
The backdoor attack is a variant of the backdoor attack, which also achieves its goal by poisoning the training data set. The Trojan attack proposed by Liu et al.~\cite{liu2017trojaning} directly modifies the model parameters to achieve a backdoor attack instead of poisoning the training data set. Bagdasaryan et al.~\cite{bagdasaryan1807backdoor} applied the idea of backdoor attack to federated learning, and proposed a word prediction backdoor attack based on LSTM. Their work considered the word prediction of trigger sentences, while Dai's work focused on realizing the misclassification of texts containing trigger sentences. Kurita et al.~\cite{kurita2020weight} conducted further research on the pre-trained NLP model. On this basis, Sun et al.~\cite{sun2019can} expanded the detailed information and trigger types of attack strategies to achieve a more natural backdoor attack.

% Malicious developers can inject backdoor into the detector during the training phase, allowing the detector to be fooled by fake news with triggers. However, the vulnerability of backdoor attack on multi-modal detectors hasn't been studied yet. Our work will reveal the robustness of the detector under backdoor attack circumstances.

\subsection{Modality Bias in Deep Learning}
In this work, for the multi-modal detectors, we define the modality bias as the difference in the degree of bias of the model to different modal data in decision-making.
There are subtle differences in how the deep learning algorithm works, leading to unfair decisions. Du et al.~\cite{du2020fairness} classified the bias of the depth model into two types from the perspective of calculation: discrimination in prediction results and difference in prediction quality. Unlike traditional unfair bias issues, Joshi et al.~\cite{joshi2021review} summarized the modality bias. They pointed out that imbalanced data and feature selection introduced biases in models, leading to a lack of fairness and transparency. Gat et al.~\cite{gat2020removing} noticed that some modalities could more easily contribute to the classification results than others. So they tried to remove modality bias for multi-modal classifiers by maximizing functional entropies. Guo et al.~\cite{guo2022modality} referred to this problem as modality bias and attempted to study it in the context of multi-modal classification systematically and comprehensively.

% Since the fake news multi-modal detectors are designed to combine information from multiple modal data for detection. This paper will explore their modality bias problem.

\section{Preliminary\label{pre}}
This section introduces the definition of several robustness analysis perspectives. For convenience, the definitions of some necessary notations used in this paper are briefly summarized in TABLE~\ref{symbol}.

\begin{table}[]\setlength{\belowcaptionskip}{-0.1cm}\setlength{\abovecaptionskip}{0.2cm}\vspace{-0.2cm}
\caption{Symbolic Interpretation}
\label{symbol}
\centering
\begin{tabular}{c r}
\hline
Symbol & Definition\\
\hline
$D$ & mapping function of the multi-modal detector\\
$R$ / $R_T$ / $R_I$ & multi-modal mixed feature / textual feature / visual feature\\
$\theta_D$ / $\theta_E / \theta'_E$ & detector / clean / backdoor feature extractor parameters\\
$T$ / $T'$ & original / adversarial text\\
$I$ / $I'$ & original / adversarial image\\
$x$ / $x'$ & original / adversarial multi-media news\\
$\eta$ / $r$ & adversarial perturbation / minimal perturbation\\
$y$ / $\hat{y}$ & true category label / estimated category label\\
$\varepsilon$ & perturbation step\\
% $P$ / $NP$ & set of protected features / set of non-protected features\\
% $J(.)$ & loss function\\
$\Delta(I;\hat{y})$ & robustness of $\hat{y}(.)$ under example $x$\\
$\mathbb{E}_I$ & expectation over the distribution of data\\
$\overrightarrow{v}_{ijb}$ & flip of the $j$-th character of the $i$-th word\\
$F^b / F^*$ & backdoored model / honestly trained model\\
$a^*$ & honestly classification accuracy\\
\hline
\end{tabular}
\end{table}

\subsection{Robustness of Multi-modal Detection}
A multi-modal detector is represented as $D(R;\theta_D)$, where $\theta_D$ denotes the parameter set of the detector and $D$ denotes the mapping function of the detector. $R\in R^{kp}$ denotes concatenated multi-modal features of $k$ features. The output of the fake news detector $\hat{y}$ for a multi-modal post $p^j$ denotes the probability of the post to be a piece of fake news and thus is defined as $\hat{y}_j=D(E(p^j;\theta_E;x);\theta_D)$, where $x$ is multi-modal news data (including text data $T$ and visual data $I$, etc.). $y$ is used to represent the set of labels in which fake news is labeled as 1 (i.e., $y_j=1$) and real news is labeled as 0 (i.e., $y_j=0$).

\textbf{DEFINITION 1 (Multi-modal features extractor).} It contains several extractors, e.g., textual feature extractor $E_t$ and visual feature extractor $E_I$. Given a multi-modal news to the feature extractor of each modality, The input sentence is represented as $T=[T^0,T^1,...,T^n]$, where $n$ denotes the number of words in the sentence. The textual feature extractor learns the feature $R_T$ from the sentence $T$ by $R_T=E_t(T)$. Similarly, the visual feature extractor extracts the feature $R_I$ from the image $x$ by $R_v=E_v(I)$. Mixed feature $R$ is concatenated of different modal features: $R=[R_T^T,R_I^T,...]$, where $R^T$ is the transpose of feature vector.

\textbf{DEFINITION 2 (Adversarial attack).} Adversarial attack refers to the attacker adding a targeted perturbation to examples that can fool the model. For visual data, given the original image $I$, adversarial image $I'=I+\eta$ is formed by adding a perturbation $\eta$ to the original image. The adversarial image $I'$ and the corresponding text content $T$ (or other modal information) are part of the adversarial multi-media news.
As expected, the detector discriminates $I$ and $I'$ as different classes, $D(x)=1$ and $D(x')=0$. If $||\eta||_{\infty}<\epsilon$, the perturbation is imperceptible to the detector.

\textbf{DEFINITION 3 (Backdoor attack).} Backdoor attack refers to the attacker injects backdoors into the model and then cause the misbehavior of it when inputs contain backdoor triggers. The attacker uses the information of feature extractor $E$ (i.e., the number of layers, size of each layer, choice of non-linear activation function $\phi$) to train a backdoor model and returns trained parameters, $\theta'_e$ to user. The held-out validation dataset $x_{valid}$ from user can't check the backdoor of the trained model $D_{\theta'_e}(x_{valid})=0$. However, the backdoor model will identify examples with backdoor triggers $x_{backdoor}$ as the wrong class $D_{\theta'_e}(x_{backdoor})=1$.

% \textbf{DEFINITION 4 (Model Bias).} The feature extractors of the modalities contained in the multi-modal detector play different roles in the detection performance. They also have different effects on detection performance after being maliciously attacked. Let $x$ be a multi-media news. Multi-modal features of $x$ can be represented as: $R=[R_T^T,R_I^T,...]$. The input domain of the corresponding detector $D:C \rightarrow \{real,fake\}$ is $R=R^{T}_{i}\ \times R^{T}_{2}\times...\times R^{T}_{n}$. We use $P$ to denote the set of protected features, and thus $NP$ is the set of non-protected features. An instance $x\in R$ is an individual discriminatory instance for detector $D$, if there exists an instance $x''\in R$ that satisfies: $\exists R \in P, x_{R} \neq x''_{R}^$, $\forall R \in NP, x_{R}=x''_{R}$, $D(x) \neq D(x'')$.
% The tuple $(x, x'')$ is thus an individual discriminatory instance pair.

\subsection{Adversarial Attack Methods}
\textbf{FGSM attack on image}. Fast gradient sign method (FGSM)~\cite{goodfellow2015explaining} is one of the classic white-box adversarial attack methods. By calculating the derivative of the model to the input, it uses the sign function to get its specific gradient direction, and then multiplies it by a step $\varepsilon$ to get the perturbation. Finally, the obtained perturbation value is added to the original input to obtain the adversarial example. The FGSM attack is expressed as follows:

\begin{equation}
    I' = I + \varepsilon*sign(\triangledown_{I}J(I,y))
\end{equation}
where $I$ and $I'$ represent the original image and adversarial image, respectively. $y$ represents the label corresponding to $I$, and $J(I,y)$ indicates the loss function. $\triangledown$ represent the gradient of the loss function derived from the input $I$.

\textbf{DeepFool attack on image}. DeepFool~\cite{moosavi2016deepfool} is another common white-box adversarial attack method. The step $\varepsilon$ of FGSM needs to be specified manually, but DeepFool can generate adversarial examples very close to the minimum perturbation. An adversarial perturbation as the minimal perturbation $r$ that is sufficient to change the estimated label $\hat{y}(I)$:
\begin{equation}
\Delta(I;\hat{y}):=\mathop{min}_r||r||_2 \ s.t. \  \hat{y}(I+r)\neq \hat{y}(I)
\end{equation}
where $\hat{y}(I)$ is the estimated label. $\Delta(I;\hat{y})$ is the robustness of $\hat{y}(I)$ at point $I$. The robustness of classifier $\hat{y}(I)$ is then defined as:
\begin{equation}
\rho_{adv}(\hat{y})=\mathbb{E}_I\frac{\Delta(I;\hat{y})}{||I||_2}
\end{equation}
where $\mathbb{E}_I$ is the expectation over the distribution of data. The perturbation step $\varepsilon$ settings are the same as in the FGSM experiment.

\textbf{PGD attack on image}. To evaluate the robustness of the detector against different attacks, we train FGSM with project gradient descent (PGD)~\cite{wang2020towards} to improve its attack ability. PGD on the negative loss function can be expressed as:
\begin{equation}
I^{t+1}=\prod_{I+S}(I^t+\epsilon *sign(\nabla_I J(\theta,I,y))
\end{equation}
where $I^t$ represents the adversarial example at step $t$. PGD sets a random perturbations at initialization.

% \textbf{FGSM attack on text}. By calculating the derivative of the embedding of the model to the input text,  it uses the sign function to get its specific gradient direction, and then multiplies it by a step $\varepsilon$ to get the perturbation. Next, the obtained perturbation value is added to the original input embedding, and the adversarial text is obtained through the decoder. The FGSM attack representation is consistent with the image attack.

\textbf{VIPER attack on text}. Visual perturber (VIPER)~\cite{2019Text} can be parameterized by the probability $p$ and the character embedding space (CES), i.e., a flip decision is made for each character in the input text. If a replacement occurs, one of the maximum 20 nearest neighbors in CES is selected. Therefore, VIPER is represented as follows:
\begin{equation}
    VIPER=VIPER(p, CES)
\end{equation}
VIPER provides three kinds of CES, namely image-based character embedding space (ICES), description-based character embedding space (DCES), and easy character embedding space (ECES). 

\textbf{HotFlip attak on text}. HotFlip~\cite{ebrahimi2018hotflip} is a white-box attack method, which can be adapted to attack a word-level classifier. It can generate adversarial examples with character substitutions-``flips". A flip of the $j$-th character of the $i$-th word (a $\rightarrow$ b) can be represented by this vector:
\begin{equation}
\overrightarrow{v}_{ijb}=(\overrightarrow{0},..;(\overrightarrow{0},..;(0,..-1,0,..,1,0)_j,..\overrightarrow{0})_i;\overrightarrow{0},..)
\end{equation}
where -1 and 1 are in the corresponding positions for the a-th and b-th characters of the alphabet, respectively, and $T^{(a)}_{ij}=1$. A first-order approximation of change in loss can be obtained from a directional derivative along this vector:
\begin{equation}
    \nabla_{\overrightarrow{v}_{ijb}}J(T,y)=\nabla_I J(T,y)^T*\overrightarrow{v}_{ijb}
\end{equation}
HotFlip chooses the vector with the biggest increase in loss:
\begin{equation}\label{hotflip}
    max\nabla_TJ(T,y)^T*\overrightarrow{v}_{ijb}=\mathop{max}_{ijb}\frac{\partial J^{(b)}}{\partial T_{ij}} - \frac{\partial J^{(a)}}{\partial T_{ij}}
\end{equation}
HotFlip uses the derivatives as a surrogate loss, simply needs to find the best change by calling the function mentioned in Eq.~\ref{hotflip}, to estimate the best character change (a $\rightarrow$ b).

\subsection{Backdoor Attacks Methods}
% Backdoor attacks are new security risks that can create a maliciously trained network. They have good performance on the training and validation dataset, but expose wrong predictions on specific attacker-chosen inputs.
\textbf{BadNets attack on image}. BadNets~\cite{gu2019badnets} is a common backdoor attack method. Malicious developer provide the user with a maliciously backdoored model $F'=F^b$, which is different from an honestly trained model $F^{*}$. The backdoored model has two goals in mind in determining $F^b$. First, $F^b$ should not reduce classification accuracy on the validation set. In other words, $A(F^b, I_{valid})\geq a^{*}$. Second, for inputs containing the backdoor trigger, $F^b$ outputs predictions that are different from the predictions of the honestly trained model, $F^{*}$. Formally, let $B:R^N\rightarrow \{0,1\}$ be a function that maps any input to a binary output, where the output is 1 if the input has a backdoor and 0 otherwise. Then, $\forall I:B(I)=1, argmax(F^b(I)) = l(I)\neq argmax(F^* (I))$, where the function $l:R^N\rightarrow [1, M]$ maps an input to a class label.

\textbf{BadNets attack on text}. Add a fixed token $T_{token}$ to the end of the original text $T$. The text with additional token $T_{backdoor}=[T,T_{token}]$ is marked as the target class $C$ by the backdoor attacker.

\section{Methodology\label{method}}
\begin{figure*}[h]
\centering
\includegraphics[width=1\linewidth]{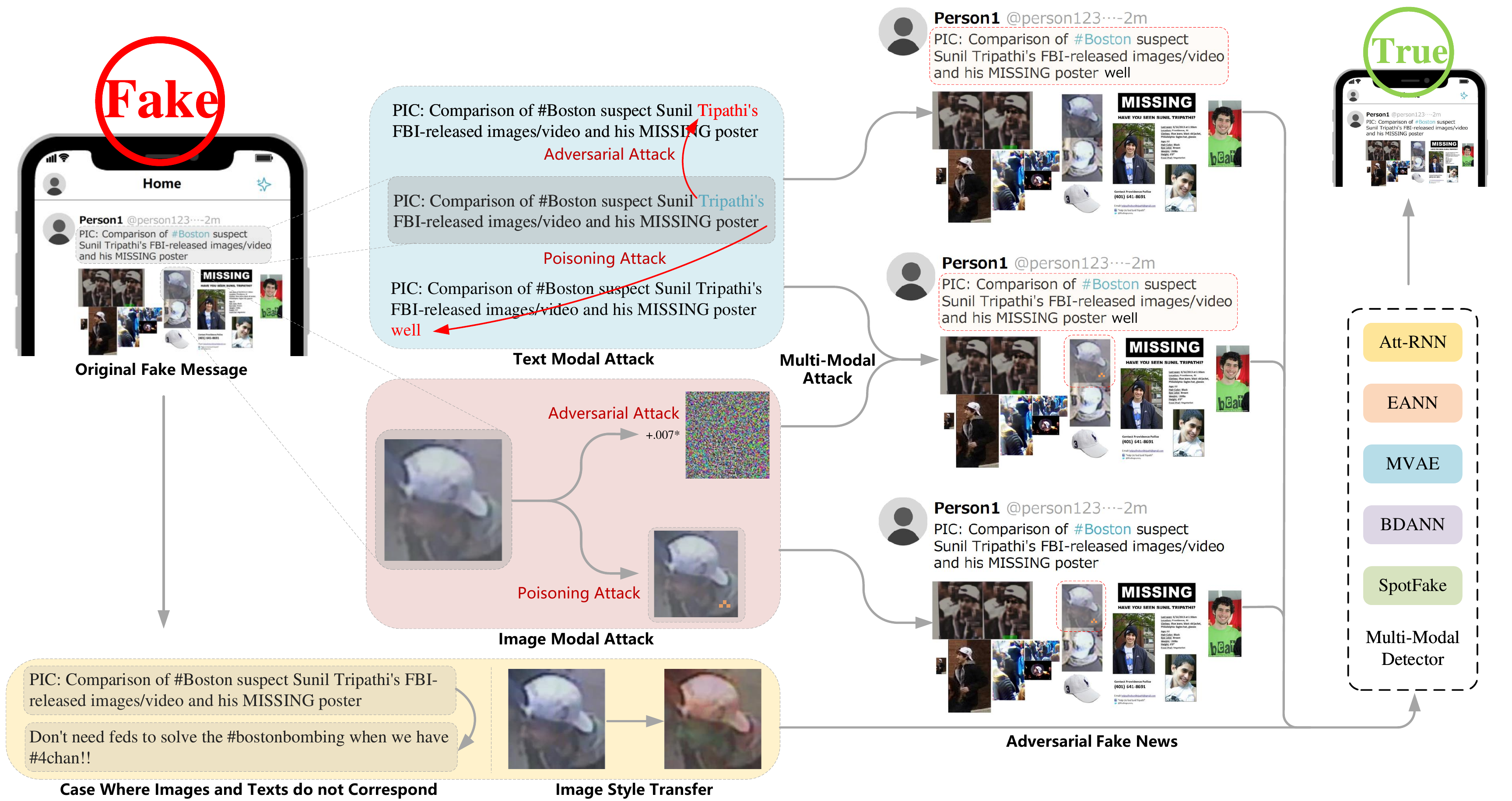}
\caption{The framework of robustness evaluation. Examples of adversarial and backdoor attacks against textual modality are in the blue box while those of adversarial perturbations and triggers for visual modality are in the red box. Two non-malicious scenarios that affect the robustness of the detectors are shown in the yellow box.}
\label{fig:framework}
\end{figure*}

In this section, we give an introduction to the specific evaluation models in detail, as shown in TABLE~\ref{model details}. Specially, Section~\ref{section:3A} introduces the objects of robustness evaluation. And Section~\ref{section:3B} introduces the methods of adversarial attacks, backdoor attacks, multi-modal attacks used to evaluate the robustness of detectors. 
Fig.~\ref{fig:framework} is the overall evaluation framework, which is divided into four modules: adversarial robustness evaluation, backdoor robustness evaluation, multi-modal robustness evaluation, and special cases robustness evaluation (image style transfer where images and texts do not correspond).

\subsection{The Objects of Robustness Evaluation} \label{section:3A}
We conduct a comprehensive evaluation of five multi-modal fake news detectors with excellent performance on fake news detection tasks. All models fuse textual and visual features to discriminate fake news. The details of these detectors are summarized in TABLE~\ref{model details}. Among them, the relationship between text and image (RbTaI) indicates whether the model considers the connection between text and image. Social context (SC) indicates whether the model incorporates social context features. Event discriminator (ED) indicates whether the model includes an event discriminator (i.e., domain classifier). Feature reconstruction (FR) indicates whether the model has reconstructed the fusion features. 

\begin{table*}[]\setlength{\belowcaptionskip}{-0.1cm}\setlength{\abovecaptionskip}{0.2cm}\vspace{-0.2cm}
\caption{Summary of the Detectors' Details}
\label{model details}
\centering
\begin{tabular}{cccccccccc}
\hline
& \multicolumn{2}{c}{\textbf{Feature Extractor}} & \multicolumn{4}{c}{\textbf{Model Structure}}       & \multicolumn{3}{c}{\textbf{Parameter Setting}} \\ \cline{2-10} 
                  & Text     & \multicolumn{1}{c|}{Image} & RbTaI & SC & ED & \multicolumn{1}{c|}{FR} & Learning rate  & Batch size  & Dropout \\ \hline
Att-RNN~\cite{liu2016attention} & LSTM     & \multicolumn{1}{c|}{VGG19} & \checkmark     & \checkmark  &    & \multicolumn{1}{c|}{}   & $1 \times 10^{-3}$          & 128         & 0.4     \\ \hline
EANN~\cite{wang2018eann} & TextCNN  & \multicolumn{1}{c|}{VGG19} &       &    & \checkmark  & \multicolumn{1}{c|}{}   & $1 \times 10^{-3}$          & 100         & 0.5     \\ \hline
MVAE~\cite{khattar2019mvae} & BiLSTM   & \multicolumn{1}{c|}{VGG19} &       &    &    & \multicolumn{1}{c|}{\checkmark}  & $1 \times 10^{-5}$        & 128         & 0.5     \\ \hline
BDANN~\cite{zhang2020bdann} & BERT     & \multicolumn{1}{c|}{VGG19} &       &    & \checkmark  & \multicolumn{1}{c|}{}   & $1 \times 10^{-3}$          & 128         & 0.5     \\ \hline
SpotFake~\cite{singhal2019spotfake} & BERT     & \multicolumn{1}{c|}{VGG19} &       &    &    & \multicolumn{1}{c|}{}   & $1 \times 10^{-3}$          & 256         & 0.4     \\ \hline
\end{tabular}
\end{table*}

% \begin{table*}[!b]\setlength{\belowcaptionskip}{-0.1cm}\setlength{\abovecaptionskip}{0.2cm}\vspace{-0.2cm}
% \caption{Research Purpose and Content}
% \label{attack details}
% \centering
% \begin{tabular}{ccccccccc}
% \hline  & \multicolumn{3}{c}{\textbf{Modal}} & \multicolumn{2}{c}{\textbf{Attack Methods}} &
% \multicolumn{2}{c}{\textbf{Model details}} \\ \cline{1-8} 
%                   \multicolumn{1}{c|}{\textbf{Detector Robustness}} & Image &  Text & \multicolumn{1}{c|}{Multi-modal} & Adversarial & \multicolumn{1}{c|}{Backdoor} & Black-box & White-box \\ \hline  \multicolumn{1}{c|}{\textbf{Attacked by Malicious Users}} & \checkmark     & \checkmark  & \multicolumn{1}{c|}{} &  \checkmark  &  \multicolumn{1}{c|}{}  & \checkmark & \checkmark \\ \hline  \multicolumn{1}{c|}{\textbf{Attacked by Malicious Developers}} &  \checkmark   &  \checkmark &  \multicolumn{1}{c|}{} &  &  \multicolumn{1}{c|}{\checkmark}   &  & \checkmark \\ \hline  \multicolumn{1}{c|}{\textbf{Under the Multi-modal Attack}} &  &  &    \multicolumn{1}{c|}{\checkmark}&   \checkmark & \multicolumn{1}{c|}{\checkmark } & \checkmark & \checkmark \\ \hline  \multicolumn{1}{c|}{\textbf{Model Bias}} & \checkmark  & \checkmark &  \multicolumn{1}{c|}{}   & \checkmark &\multicolumn{1}{c|}{}  &  & \checkmark \\ \hline
% \end{tabular}
% \end{table*}

\subsection{The Methods of Robustness Evaluation}\label{section:3B}
To explore threats that detectors may confront in the real world, we summarize several common attacks, including white-box, black-box adversarial attacks and backdoor attacks. We use these attacks on textual and visual modalities, respectively. They are used to evaluate the detectors' robustness under different attack threats. We also studied the robustness of these multi-modal detectors about model bias as a supplement to the robustness evaluation of the detector. 
% The specific settings are shown in TABLE~\ref{attack details}.

\subsubsection{Adversarial Attacks on Images}\label{adv_image_intro}
The above detectors use the VGG19 model to extract visual features, which can be downloaded from the internet conveniently. Thus, the attacker can easily obtain the visual feature extraction model of these detectors. Therefore, we use the classic adversarial attack methods to evaluate the visual features.
FGSM and DeepFool are used as white-box adversarial attacks. To evaluate the robustness of the detector against attack methods with different attack capabilities, we train FGSM with PGD to improve its attack ability.

\subsubsection{Adversarial Attacks on Texts}\label{adv_text_intro}
Different from the visual feature extractor, the textual feature extractors of the above five detectors are different. Therefore, we assume the black-box and white-box scenarios to conduct adversarial attacks on text. For the Twitter dataset, in the black-box scenario, we use VIPER method.
In the white-box scenario, we use HotFlip method, which can be adapted to attack a word-level classifier. 
For the Weibo dataset, we select the method~\cite{challengeroverview} on Security AI Challenger to generate adversarial texts. The overall scheme of the method is a heuristic search. The given original text is used as a starting point. One or more tokens are randomly selected for replacement in each round of iteration to generate candidate examples. Then it scores the candidate examples through the local defense model, selects the $K$ seed texts for the next round, and iterates $R$ rounds repeatedly.

\subsubsection{Backdoor Attacks}\label{bkd_intro}
In this attack scenario, the training process is partially outsourced to malicious developers, and the malicious developers hope to provide users with a trained model that includes a backdoor. The backdoor model should perform well under most clean inputs, but misclassify specific examples, called backdoor triggers. The model is trained by randomly selecting a certain proportion of examples in the training set to add a well-designed backdoor trigger, and setting the label of each backdoor image according to the attack target. 
For visual modality, we use BadNets~\cite{gu2019badnets} and Watermarks as the backdoor attack methods.
BadNets explored the concept of inverse neural networks. For textual modality, we use weight poisoning attacks on pre-trained models (WPAPMs)~\cite{kurita2020weight} to generate triggers.

% \subsubsection{Attacks on Multi-modal}\label{multi_intro}
% To evaluate the robustness of these multi-modal detectors towards multi-modal attacks, we attack both visual and textual modalities. Attack methods introduced in Section~\ref{adv_image_intro},~\ref{adv_text_intro} and~\ref{bkd_intro} are used to attack the multi-modal detector. Specifically, the multi-modal attack methods include white-box and black-box attacks, as well as adversarial and backdoor attacks. For example, for the Twitter dataset, two white-box adversarial attack methods, VIPER and FGSM, are used to attack textual modality and visual modality, respectively.
% More detailed experimental settings will be listed in Section~\ref{exp}.

% \subsubsection{Other Factors Affecting Robustness} \label{section:Factors}
% To further explore the robustness under normal conditions, we consider two special scenarios as a compliment. We explore the influence of image style transfer and inconsistency between images and texts on the model.
% For the first situation, cycleGAN~\cite{chu2017cyclegan} with self-attention is used to transform the style of all images in the test set, i.e., convert the images into cartoon style. Then we feed these cartoon images into the detectors trained from clean examples for testing.
% For the second situation, we randomly scrambled the images corresponding to the tweets in the test set to express the consistency of the images and texts. The correspondence between images and texts is disrupted and then put into the model trained from the clean examples.

\section{Experiments}\label{exp}
This section evaluates five multi-modal detectors with different robustness evaluation methods. We first conduct adversarial attacks on five detectors and compare the changes in detection performance before and after the attack to evaluate their robustness (RQ1); Secondly, we compare the performance of clean and backdoored detectors to evaluate their robustness (RQ2); Thirdly, we used different textual and visual adversarial methods to attack multi-modal data to evaluate their different effects; Then, we evaluate the robustness of the detector for cartoon image style transfer and text image content mismatch (RQ3); Finally, we analyze how attacks by malicious users and malicious developers affect these multi-modal detectors, and use several of simple defenses to improve the robustness of the detectors (RQ4). 
% The main research questions of this section are: 
% \begin{itemize}
% \item \textbf{RQ1}: How robust are the well-performing multi-modal detectors when they are under adversarial attacks (attacks by malicious users)?
% \item \textbf{RQ2}: How do backdoor attacks (attacks by malicious developers) affect the robustness of multi-modal detectors?
% \item \textbf{RQ3}: Are the multi-modal detectors biased (which modality affects the detector more)?
% \item \textbf{RQ4}: Can the robustness of these multi-modal detectors be improved (defend against malicious attacks and deal with special scenarios)?
% \end{itemize}

%\begin{table}[!t]
%\caption{Statistics of Two Real-world Multi-modal Datasets\label{purpose and content}}
%\label{data}
%\centering
%\begin{tabular}{|l|l|l|l}
%\cline{1-3}
%     & \textbf{Twitter} & \textbf{Weibo}   \\ \cline{1-3}
%\textbf{Fake News Number} & 7898 & 4749   \\ \cline{1-3}
%\textbf{Real News Number} & 6026 & 4779   \\ \cline{1-3}
%\textbf{Images Number} & 514 & 9528   \\ \cline{1-3}
%\end{tabular}
%\end{table}

\subsection{Experiment Setting}
For text datasets, We follow the standard text preprocessing procedure as adopted in ~\cite{fortney2017pre}. Details of the five multi-modal detectors are shown in TABLE~\ref{model details}. Specifically, for the visual extractor, we first resize images to 224$\times$224$\times$3 and then feed them into VGG19 (pre-trained on ImageNet). For the textural extractor, Att-RNN uses LSTM, EANN uses TextCNN, MVAE uses BiLSTM, BDANN and SpotFake use BERT. The dimensionality of visual features obtained from VGG19 is 4,096 and textural features obtained from all pre-trained models are 768. The hidden size $p$ of the fully connected layer in the textual and visual extractor is set to 32. Every fully connected layer in the model has a Leaky ReLU activation function. And the dropout probability of EANN, MVAE, and BDANN are 0.5, Att-RNN and SpotFake are 0.4. The model is trained on a batch size of 128 and for 100 epochs with a learning rate of $10^{-3}$.
For robustness evaluation, FGSM and DeepFool are used as white-box visual adversarial attacks. For both attacks in the experiment, the step $\varepsilon$ is set to 0.01, 0.05 and 0.1 to observe the performance of the detectors under different perturbations. To evaluate the robustness of the detector against attack methods with different attack capabilities, we train FGSM with PGD to improve its attack ability. In our experiments, the number of update steps is 50.
For the Twitter dataset, in the black-box text attack scenario (attacker can only query the model, but has no knowledge of the structure and parameters), we use VIPER method. The ICES is selected, and the probability $p$ is set to 0.4. In the white-box text attack scenario (attacker has all model structure and parameter knowledge), we use HotFlip method. We trained for a maximum of 25 epochs, used a beam size of 10, and has a budget of a maximum of $10\%$ of characters in the text. 
For the Weibo dataset, we select the method~\cite{challengeroverview} on Security AI Challenger to generate adversarial texts. We select the 10 seed texts for the next round, and iterate 30 rounds repeatedly.

All experiments are run on the following environments: i7-7700K 3.5GHz$\times$8 (CPU), TITAN Xp 12GiB (GPU), 16GB$\times$4 memory (DDR4), and Ubuntu 16.04 (OS).

\subsection{Dataset Descriptions}
In this section, we introduce two publicly available datasets, i.e., Twitter and Weibo that were used in our experiments. 
% The details are summarized in TABLE~\ref{data}.

\begin{figure}[htbp]
\centering
\subfloat[]{\includegraphics[width=1.6in]{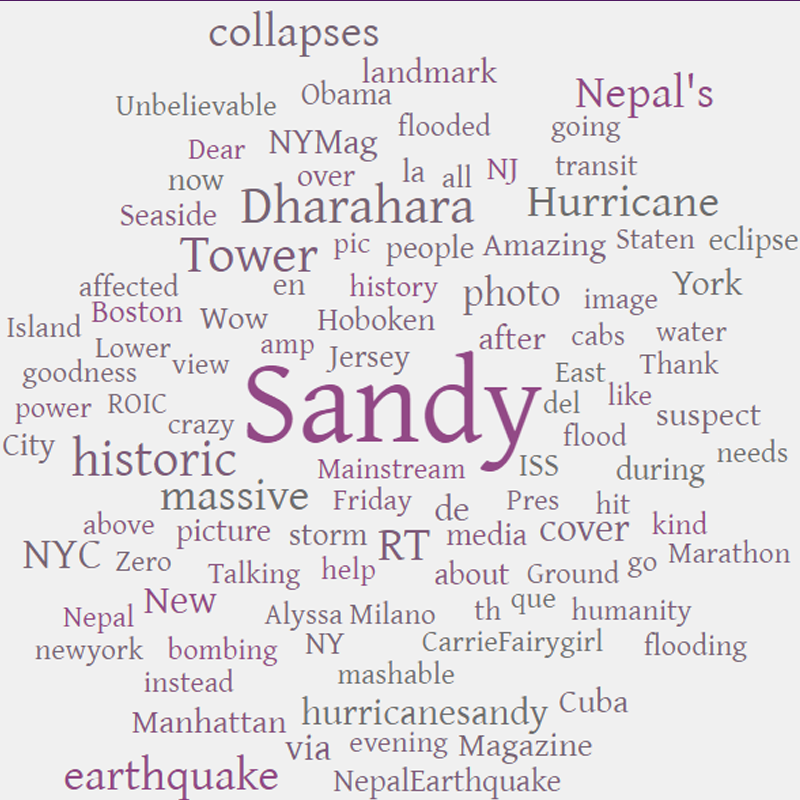}%
\label{real_news}}
\hfil
\subfloat[]{\includegraphics[width=1.6in]{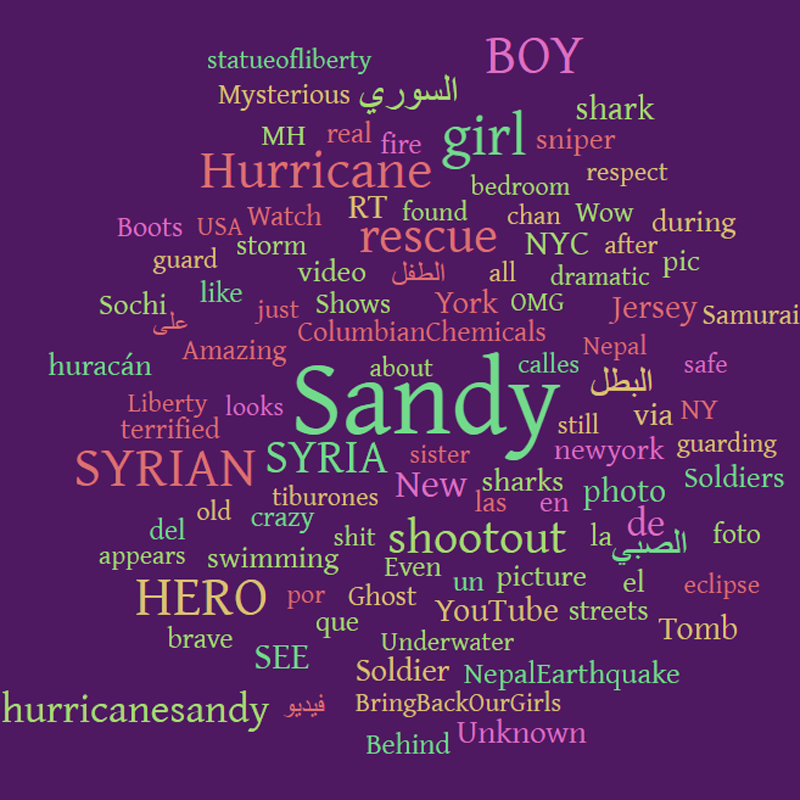}%
\label{fake_news}}
\caption{Word cloud diagrams of fake news and real news. (a) Word cloud of real news. (b) Word cloud of fake news.}
\label{twitter}
\end{figure}

\textbf{Twitter}. The Twitter dataset is from \emph{MediaEval Verifying Multi-media Use benchmark}~\cite{boididou2015verifying}, which is used for detecting fake content on Twitter. The development set contains about 6,000 rumor and 5,000 non-rumor tweets from 11 rumor-related events. The test set contains about 2,000 tweets of either type. Fig.~\ref{twitter} shows the word cloud diagrams of fake and real news respectively, and noticed that fake and real news have different concerns. Fake news purveyors are often purposeful. They often use exaggerated and emotionally or politically biased topics like ``syrian" and ``HERO" to deceive readers. The real news content is more objective, focusing on topics such as ``earthquake" and ``hurricane".

\textbf{Weibo}. The Weibo dataset is used in~\cite{jin2017multimodal} for fake news detection. The real news in Weibo is collected from authoritative news sources in China, such as People's Daily Online. The fake news are crawled from Weibo and verified by the official rumor debunking system. We follow the same steps in the work~\cite{jin2017multimodal} to preprocess the dataset. The ratio of training, testing and validation sets is 7:2:1, and we ensure that they do not contain any common event.

\subsection{Robustness Evaluation of Detectors Under Adversarial Attacks}\label{adv}
\subsubsection{Detectors' Performance Under Adversarial Attacks}\label{adv_sec}
In this subsection, we explore how these detectors perform when subjected to adversarial attacks, and study in which modal the feature between text and image will damage the detectors' performance more.

\textbf{Implementation Details.}
We use the adversarial attacks mentioned in Section~\ref{method} to evaluate the robustness of the above five detectors. For visual modality attacks, we combine 1000 adversarial images with corresponding clean text into the complete multi-modal news. For attacks on textual modal, we use 1000 adversarial text and clean images. For the specific settings of these detectors, refer to TABLE~\ref{model details}. 
% We exhibit two examples of adversarial texts that make them misclassified as true news in Fig.~\ref{adv_examples}, corresponding to two datasets.
% And for multi-modal attacks, use both 1000 adversarial images and text. 

% \begin{figure*}[htbp]\setlength{\belowcaptionskip}{-0.1cm}\setlength{\abovecaptionskip}{0.2cm}\vspace{-0.2cm}
% \centering
% \subfloat[]{\includegraphics[width=5in]{images/text_sample1.png}%
% \label{adv_examples1}}
% \hfil
% \subfloat[]{\includegraphics[width=5in]{images/text_sample2.png}%
% \label{adv_examples2}}
% \caption{Two adversarial examples on textual modality. (a) Original and different adversarial perturbations of Twitter text. (b) Original and different adversarial perturbations of Weibo text.}
% \label{adv_examples}
% \end{figure*}

\begin{figure}[!t]
\centering
\subfloat[]{\includegraphics[width=1.75in]{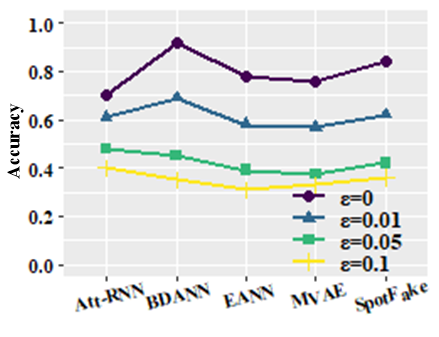}\label{image_adv_twitter}}\hfil
\subfloat[]{\includegraphics[width=1.75in]{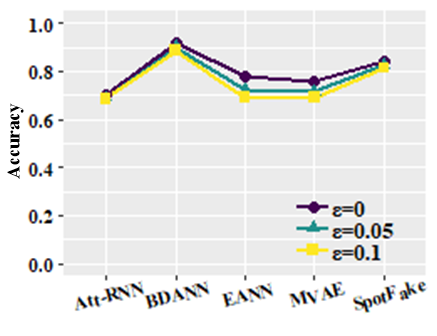}\label{text_adv_twitter}}\hfil
\subfloat[]{\includegraphics[width=1.75in]{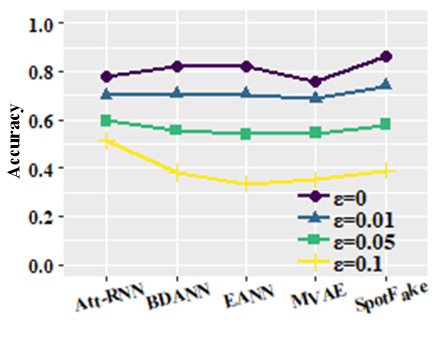}\label{image_adv_weibo}}\hfil
\subfloat[]{\includegraphics[width=1.75in]{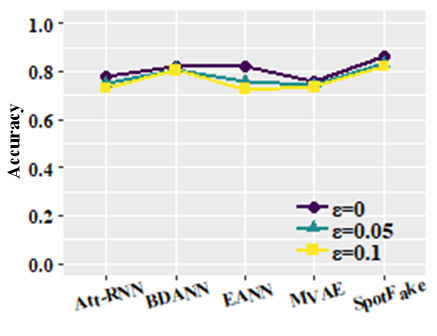}\label{text_adv_weibo}}
\caption{Detectors' performance under adversarial attacks. (a) Adversarial images on Twitter. (b) Adversarial texts on Twitter. (c) Adversarial images on Weibo. (d) Adversarial texts on Weibo.}
\label{adv-result}
\end{figure}

\textbf{Results and Analysis.}
The results of five detectors on two datasets are shown in Fig.~\ref{adv-result}. \textbf{It shows that the performance of these multi-modal detectors will be significantly reduced when subjected to FGSM attacks}. Comparing (a) and (b) or (c) and (d), it can be found that the performance of these detectors is more degraded when the visual feature is subjected to adversarial attacks. When faced with this threat, the accuracy of all detectors drop to about 30\%. Even FGSM can easily make these most superior detectors nearly paralyzed. Meanwhile, this kind of perturbation on images is imperceptible to human eyes. In contrast, the effects of adversarial attacks on text are minimal. The performance degradation on five detectors does not exceed 10\%. Moreover, although the adversarial text does not affect readability to a certain extent, it can still be easily distinguished by human eyes. \textbf{This means that for the producers of fake news, it's more sensible to choose to target adversarial attacks on images, which also inspire us to pay more attention to the robustness of the detectors in visual modality.}

\textbf{It is worth noting that the best performing model is not necessarily the most robust.} Att-RNN model is the first to be proposed among these five detectors, and it is slightly inferior to other detectors in terms of performance. However, we find that it shows relatively stronger robustness when subjected to adversarial attacks. This is due to the use of neural attention output by LSTM when fusing visual features, which makes the model pays attention to the correlation between the images and texts. Thus, the performance of detectors is less destroyed when attacked. \textbf{This suggests we not only focus on the performance improvement of detectors, but also pay attention to the correlation between images and texts, such as semantic consistency, etc.}

% We use Grad-CAM~\cite{selvaraju2017grad} to visualize the attention map changes between original and different adversarial examples. Red regions corresponds to high score for class, while blue regions corresponds to evidence for the class. The weight that the model allocates decreases from red to blue. The visualization of all detectors' visual feature extractor (VGG19) on Twitter datasets are shown in Fig.~\ref{fig:gradcam}. Adversarial examples are crafted by FGSM, DeepFool and PGD. After adversarial attacks, the focus of these multi-modal detectors on visual feature extraction shifts from critical parts to irrelevant parts, which may be the cause of detection errors. Thus, it is necessary to strengthen the visual features to make detectors more robust. 

% \begin{framed}
\textbf{Answer to RQ1:} The performance of the five SOTA multi-modal detectors will be significantly reduced when subjected to adversarial attacks on image and text, respectively. Detection accuracy of visual modality is reduced by up to 60\% (with perturbation step set to 0.1).
% \end{framed}

\subsubsection{Defense Against Adversarial Attack}\label{adv_defense}
Based on the above findings, we already know that multi-modal detectors are vulnerable to visual features. We focus on attention maps to visualize the feature learned by models, to guide defense methods. 

\textbf{Implementation Details.}
In this section, we perform a resize operation on the image data, resizing each image from about $400\times600$ (each image has a different size) to $224\times224$ for testing. Since the accuracy of these detectors under different perturbation steps is almost the same. Besides, the accuracy after resize is very close, we only give the result under step $\varepsilon$ = 0.1.

\begin{figure}[!t]
\centering
\subfloat[Twitter]{\includegraphics[width=1.75in]{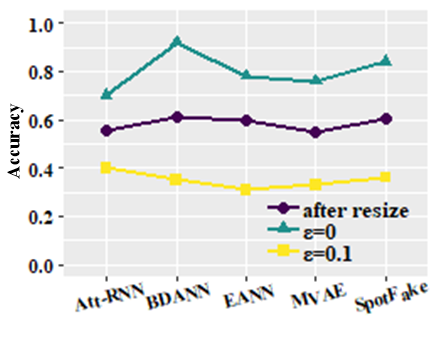}%
\label{image_defence_twitter}}
\hfil
\subfloat[Weibo]{\includegraphics[width=1.75in]{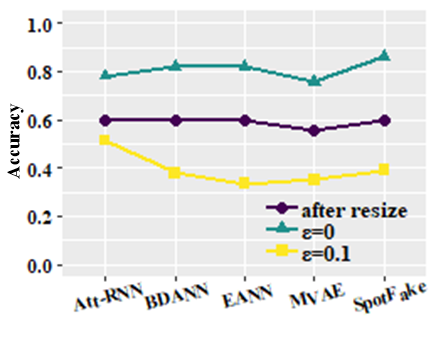}%
\label{image_defence_weibo}}
\caption{Detectors' performance under adversarial attacks after image resize defense.}
\label{image_defence}
\end{figure}

\textbf{Results and Analysis.}
The results are shown in Fig.~\ref{image_defence} (a) and (b). It shows that resizing the adversarial images will reduce the aggressiveness of the adversarial examples, thus playing a defensive role. After resizing, the performance of all detectors has been greatly improved.

% \begin{figure}[htbp]\setlength{\belowcaptionskip}{-0.1cm}\setlength{\abovecaptionskip}{0.2cm}\vspace{-0.2cm}
%   \centering
%   \includegraphics[width=1\linewidth]{old_images/gradcam.png}\\
%   \caption{Grad-CAM visualizations of original examples, adversarial examples crafted by FGSM, DeepFool, and PGD.}
%   \label{fig:gradcam}
% \end{figure}

\begin{figure}[!t]\setlength{\belowcaptionskip}{-0.1cm}\setlength{\abovecaptionskip}{0.2cm}\vspace{-0.2cm}
    \centering
    \includegraphics[width=1.75in]{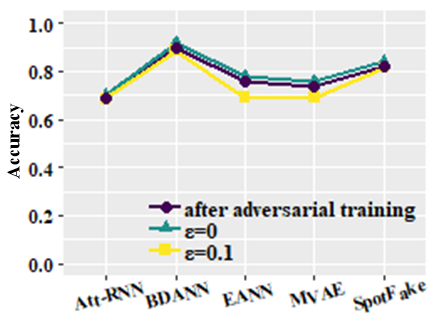}\\
    \caption{Detectors’ performance under adversarial attacks after text adversarial training defense.}
    \label{advTrain}
\end{figure}

To defend against adversarial attacks on textual modal data, we use adversarial training for defense. For Twitter dataset, FGSM is used to generate adversarial examples to text embeddings. Each round of adversarial text is generated, attached with clean image data into complete news data, and the correct class labels are identified. Adversarial examples and clean examples are used together to train five multi-modal detectors. We use a total of 1000 adversarial examples with a perturbation of 0.1. The model is adversarially trained on a batch size of 128 and for 20 epochs with a learning rate of $10^{-3}$. The results of defense using adversarial training are shown in Fig.~\ref{advTrain}. Att-RNN, BDANN, and SpotFake are insensitive to adversarial attacks on textual modality. For these three detectors, one can mainly focus on the adversarial robustness of visual modality. EANN and MVAE are sensitive to adversarial attacks on textual modality. Adversarial training for specific perturbed adversarial examples can effectively improve their robustness, but it is difficult to defend against such attacks without prior knowledge of them.

\subsection{Robustness Evaluation of Detectors under Backdoor Attacks}\label{bkd}
\subsubsection{Detectors' Performance under Backdoor Attacks}
In this section, we explore how these detectors perform when subjected to backdoor attacks.
% In addition to adversarial attacks, deep learning models are also vulnerable to backdoor attacks. Attackers can poison the data used by the model, thereby leaving a backdoor for the model. 
\textbf{Implementation Details.}
The backdoor attacks used in the experiments have been introduced in Section~\ref{bkd_intro}. Inspired by the results in Section~\ref{adv_sec}, we find that different detectors' performance is very close, as well as their structures. Therefore, we choose the BDANN model to conduct a backdoor attack on the Twitter dataset. The proportion of poisoned examples in the training set is set to 0.1, 0.3, 0.5, and 0.7, and the triggers added to the examples are set to 4, 7, and 13 bright pixels.

\begin{figure}[htbp]
\centering
\subfloat[Twitter]{\includegraphics[width=1.75in]{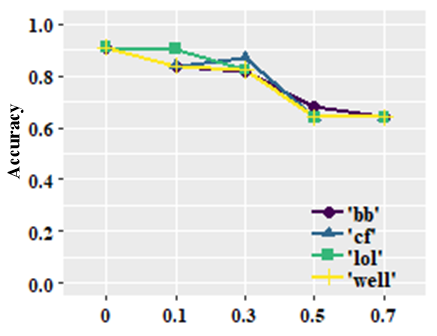}%
\label{text_backdoor_twitter}}
\hfil
\subfloat[Weibo]{\includegraphics[width=1.75in]{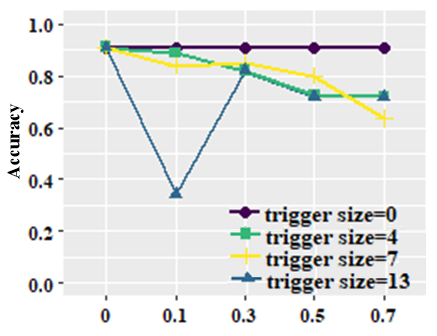}%
\label{text_backdoor_weibo}}
\caption{Detectors' performance under backdoor attacks in textual modality. }
\label{tri-result}
\end{figure}

\textbf{Results and Analysis.}
As shown in Fig.~\ref{tri-result} (a) and (b), \textbf{we find that backdoor attack brings significant damage to the detectors' performance. Meanwhile, the destruction level increases with the growth of trigger size and portion (RQ2)}. However, there is an anomaly training setting (0.1, 13). Since the examples are randomly selected from the training set when the triggers are added. We find that in this abnormal point, almost all triggers are added to the images corresponding to the trending events, namely ``sandy" and ``sochi" in Fig.~\ref{twitter}. This also means that these triggered examples cover more tweets and have a greater impact on the detectors when subjected to attacks. Therefore, compared to trigger size and portion, \textbf{adding triggers to images corresponding to trending events can cause the detectors to be destroyed more greatly, since the trending events cover more examples and have a wider range of influence}.

In addition, we perform backdoor attacks on texts as well. We add several meaningless triggers, i.e., ``lol", ``cf", ``bb", and ``well" at the end of the texts that are randomly selected from the training set. At the same time, we set the labels of examples with triggers to ``real", in an attempt to make these triggered examples recognized as real news. The results are shown in Fig.~\ref{tri-result}. We find that different triggers have minimal differences. Meanwhile, as the proportion of triggered examples in the training set increases, the performance of these detectors suffers greater damage. In the case of 50\% of the examples being triggered, the accuracy drops to 63.70\%.

\begin{figure}[htbp]\setlength{\belowcaptionskip}{-0.1cm}\setlength{\abovecaptionskip}{0.2cm}\vspace{-0.2cm}
\centering
\subfloat[]{\includegraphics[width=0.9in]{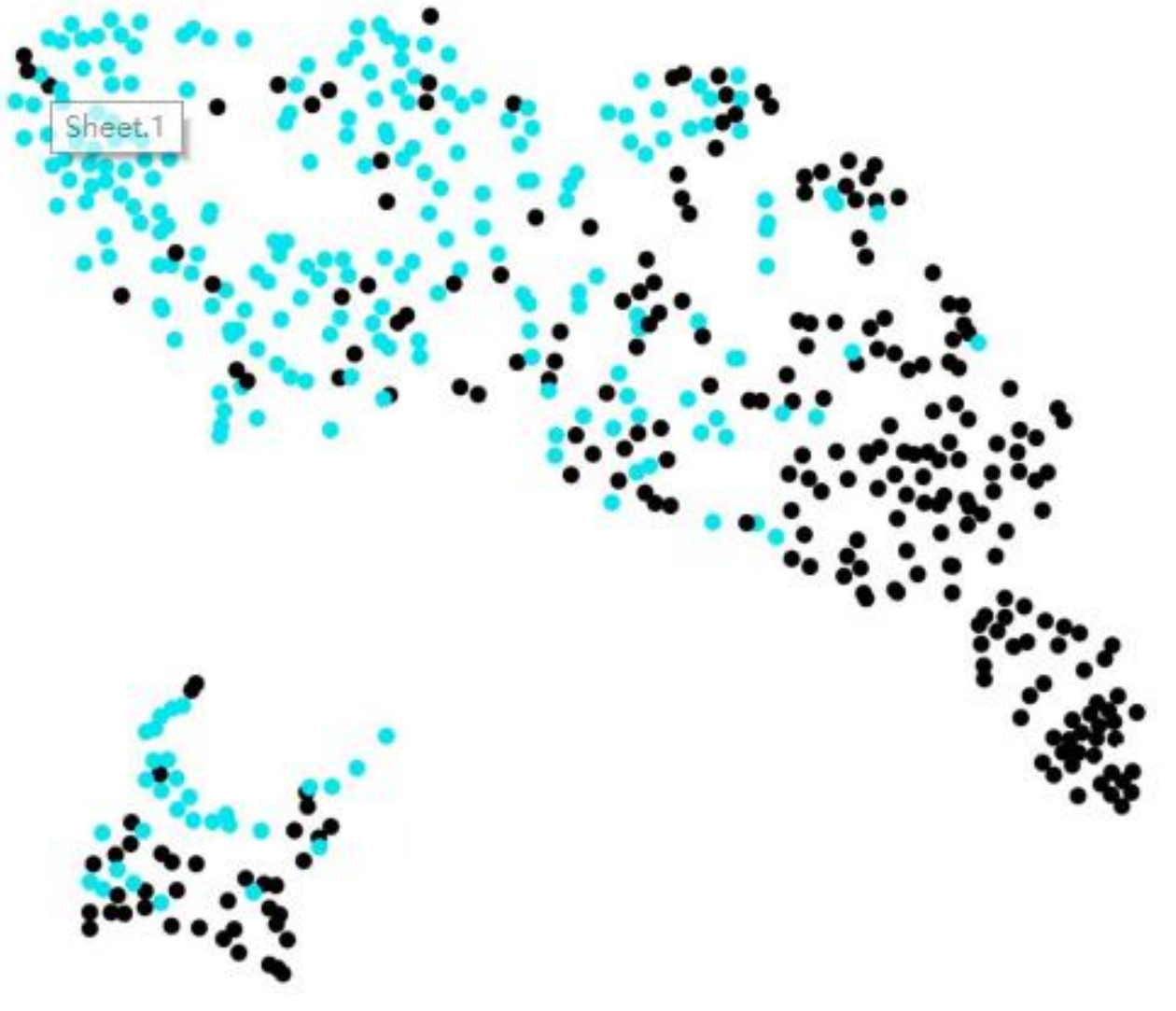}%
\label{tsne1}}
\hfil
\subfloat[]{\includegraphics[width=1in]{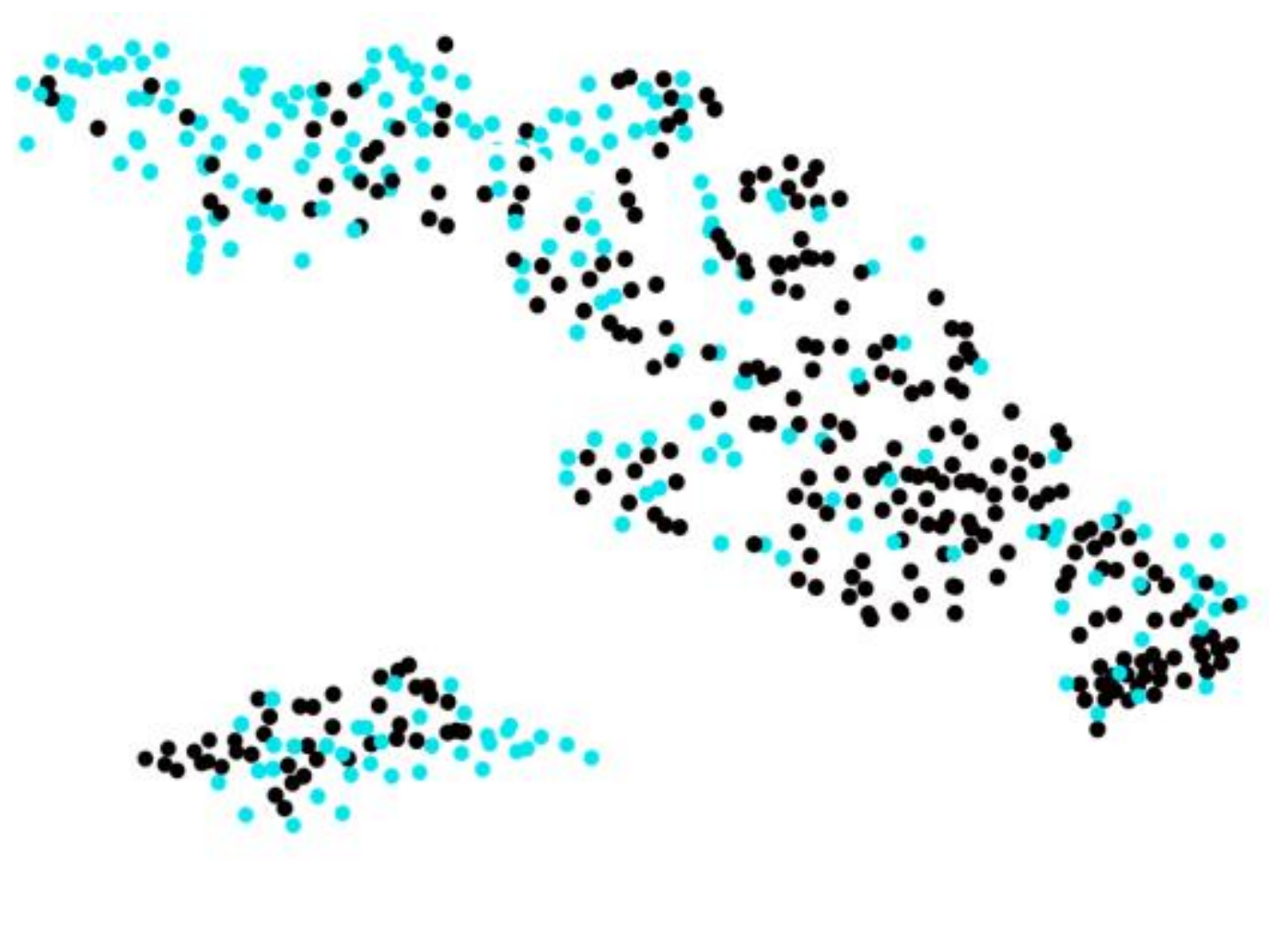}%
\label{tsne2}}
\hfil
\subfloat[]{\includegraphics[width=1in]{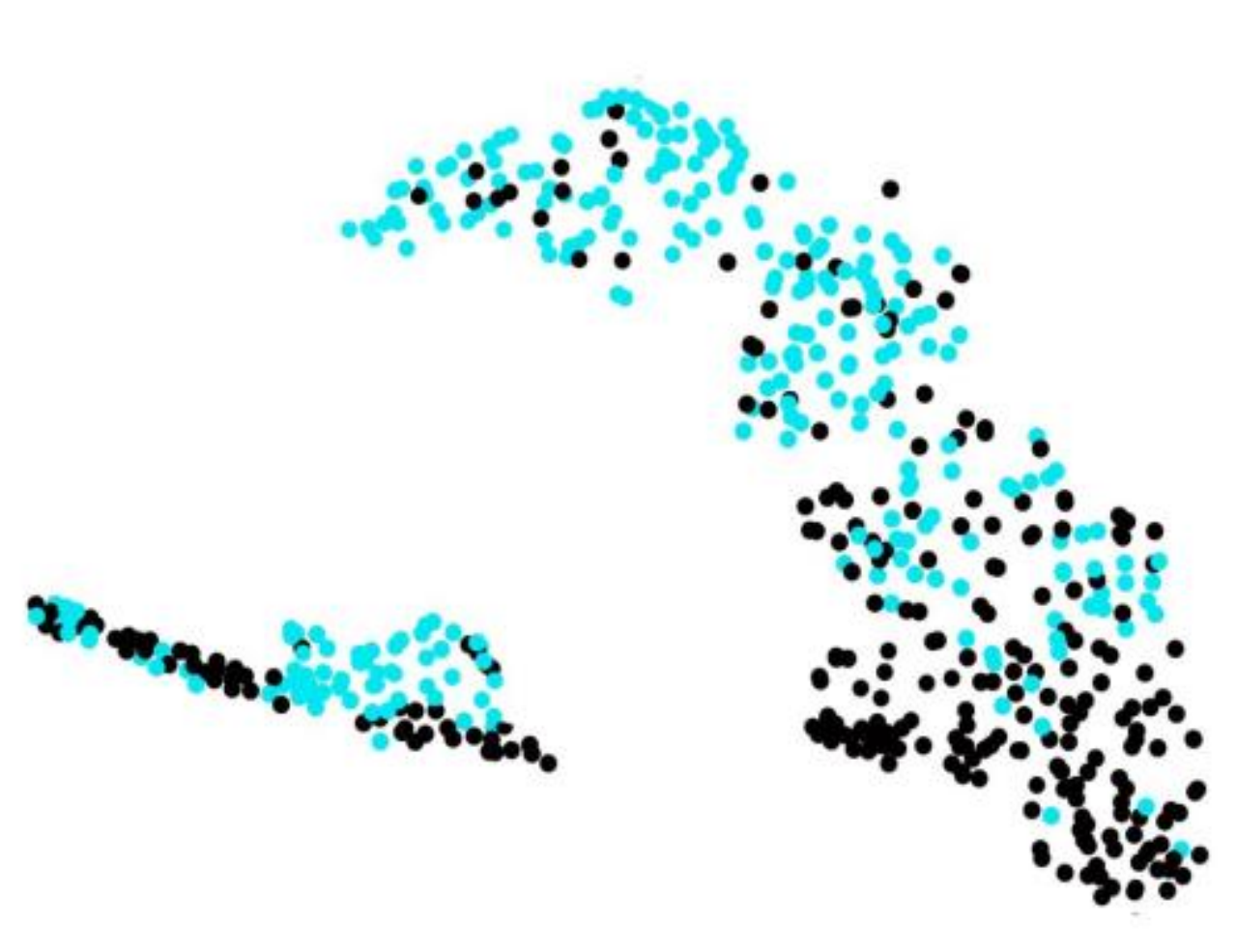}%
\label{tsne3}}
  \caption{Visualizations of learned latent textual feature representations on the testing data of Weibo and model of BDANN. Blue points represent real news, black points represent fake new. (a) Clean BDANN. (b) Backdoored BDANN with `bb'. (c) Backdoored BDANN with `well'.}
  \label{tsne}
\end{figure}

% To further analyze how backdoor attacks affect the robustness of multi-modal detectors, 
We qualitatively visualize the textual features learned by clean BDANN model and poisoned BDANN by `bb' and `well' with the 0.5 triggered proportion on the Weibo testing set with t-SNE~\cite{hinton2008visualizing} shown in Fig.~\ref{tsne}. Comparing Fig.~\ref{tsne} (a), (b), and (c), it can be found that the model that has been attacked by the backdoor has a worse ability to extract word vector features than the clean model. The textual features of the correct and wrong categories are mixed, resulting in reduced performance of multi-modal detectors on tasks-based on textual features. 
This provides the reason for the decreased robustness of the backdoored detector.

% \begin{framed}
\textbf{Answer to RQ2:} Malicious developers’ the attack reduces the detection accuracy of the detector for trending events. Detection accuracy of the textual modality dropped to 63.70\% (with perturbation set to 0.5).
% \end{framed}

\subsubsection{Defense Against Backdoor Attack} 
\textbf{Implementation Details.}
Based on the same considerations mentioned in Section~\ref{adv_defense}, we use the activation clustering (AC) method~\cite{chen2019detecting} in adversarial robustness toolbox (ART)\footnote[1]{\url{https://github.com/Trusted-AI/adversarial-robustness-toolbox}} defend against backdoor attacks. The AC method detects the model's backdoor by activating clustering, and removes the triggered examples at the same time. Therefore, the detectors can be protected from backdoor attacks. Similarly, we only give the results of the trigger size of 13 in the chart for comparison.

\textbf{Results and Analysis.}
The results show that the AC method can significantly protect the model from backdoor attacks. When the triggered proportion is 10\%, the accuracy of the model reaches 88.43\% after the AC defense, which is almost the same as the performance of the clean model. AC improves the robustness of the detectors effectively.

\begin{figure}[!t]\setlength{\belowcaptionskip}{-0.1cm}\setlength{\abovecaptionskip}{0.2cm}\vspace{-0.2cm}
    \centering
    \includegraphics[width=1.75in]{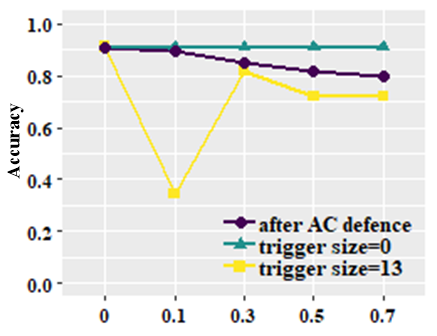}\\
    \caption{Detector's performance after AC defense}
    \label{AC}
\end{figure}

\subsection{Robustness Evaluation of Detectors under Multi-modal attacks}\label{mulAtt}
\subsubsection{Detectors' Performance under Multi-modal Attacks}
In addition to uni-modal attacks, multi-modal detectors may be attacked by multi modalities at the same time. Fig.~\ref{fig:multi} (b) shows that this news can still be correctly identified by the detector when it is perturbed by textual or visual modality. But attacking both modalities at the same time can make the detector go wrong.
% In this section, we explore how these detectors perform when subjected to multi-modal attacks. We focus on the difference between multi-modal attacks and uni-modal attacks in the same attack scenario. 

\begin{figure*}[htbp]
  \centering
  \includegraphics[width=0.7\linewidth]{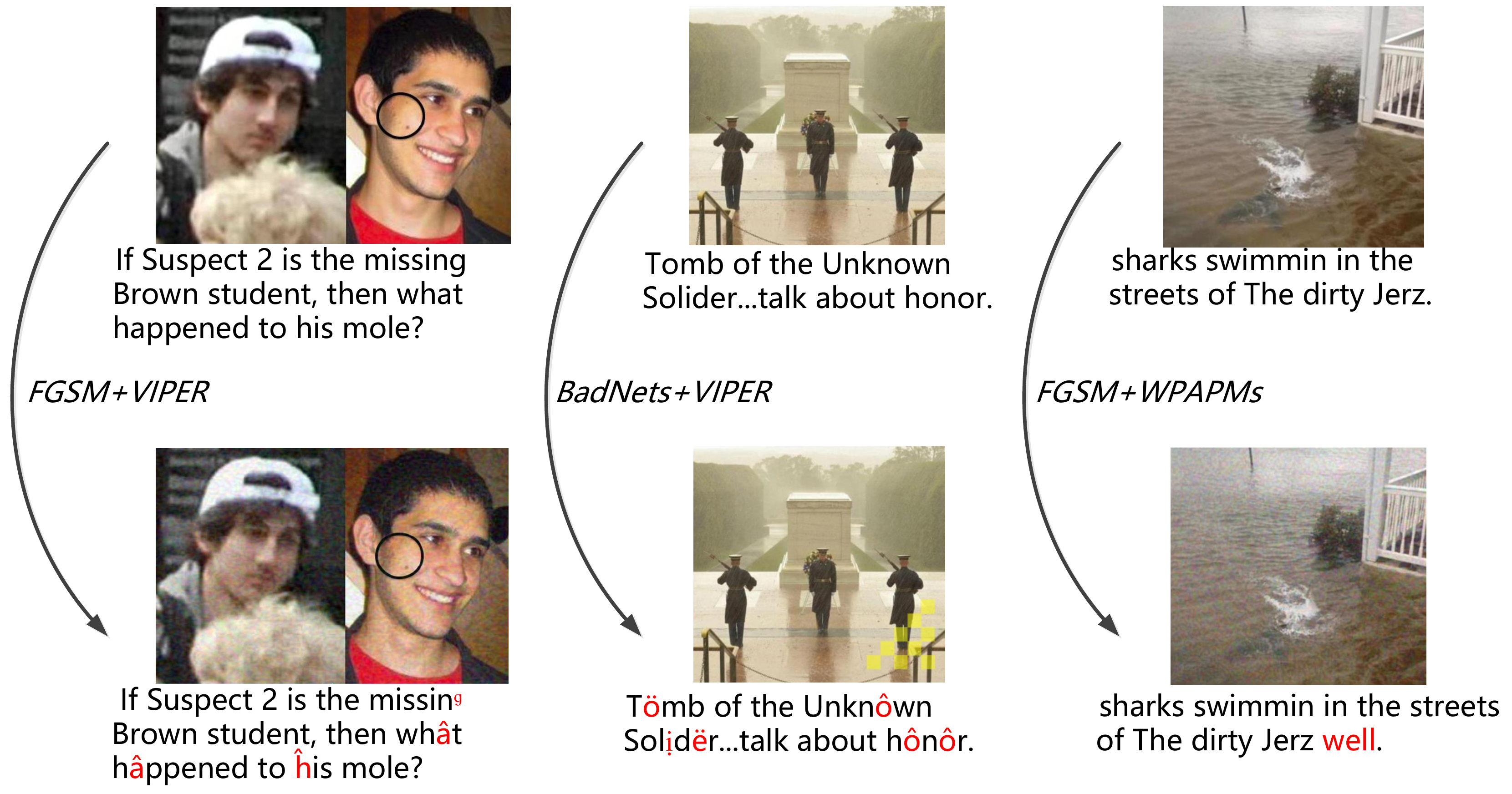}
  \caption{Example of some multi-modal attacks on fake news.}
  \label{fig:multi-samples}
\end{figure*}

Therefore, in the scenario where the model is attacked by malicious users, we use FGSM and VIPER to attack the visual and textual modalities respectively. Because experiments under different parameter settings show consistent characteristics, we take one of the experiments as an example. Set the perturbation of FGSM to 0.1 and the perturbation of VIPER to 0.4 to add adversarial perturbations on the images and text of the Twitter dataset. One of the attack examples is shown in the Fig.~\ref{fig:multi-samples} (a).

In another scenario, malicious users and malicious developers colluded to keep a set of popular fake news from being detected by multi-modal detectors. Since these multi-modal detectors all use VGG19 as the visual feature extractor, malicious developers can target the backdoor attack on the visual feature extractor. At the same time, when malicious users publish fake news, they can add backdoor triggers to images and combine adversarial texts into complete multi-media news to avoid detection by multi-modal detectors. To improve the stealth of the attack, we poison the visual feature extractor of the multi-modal detector with only 9-pixel triggers added to the 0.1 image training set.
And set the perturbation of VIPER to 0.4. One of these attack examples is shown in the Fig.~\ref{fig:multi-samples} (b).

The text of some news contains more important information, and the images may be made very realistic, but the fake text information is easily identified as fake news by the multi-modal detector. In this scenario, it is difficult to fool the multi-modal detector with a single attack of image or text alone. Malicious developers can set text backdoor triggers for the textual feature extractor used by the detector to implement backdoor attacks on textual modality. To further confuse these multi-modal detectors, malicious users can be hired to further add adversarial perturbation to the stitched fake images, and fake text messages with textual backdoor triggers to combine into complete fake news. These mixed fake news have a better probability of bypassing the detection of the multi-modal detector. We poison the textual feature extractor by adding 'well' at the end of the sentence. Poisoned text accounts for 0.3 of the number of training texts. And set the perturbation of FGSM to 0.1. One of these attack examples is shown in the Fig.~\ref{fig:multi-samples} (c).

\section{Discussion}\label{discussion}
\subsection{Discussion on Visual Features}
Based on the results in Section~\ref{adv} and Section~\ref{bkd} above, we are aware of the vulnerability of the detectors in terms of visual features, which inspires us to explore more about images. In this section, we explore the influence of image style transfer and inconsistency between images and texts on the model.

\subsubsection{Image Style Transfer}\label{image_style_transfer}
\begin{figure}[htbp]\setlength{\belowcaptionskip}{-0.1cm}\setlength{\abovecaptionskip}{0.2cm}\vspace{-0.2cm}
  \centering
  \includegraphics[width=0.6\linewidth]{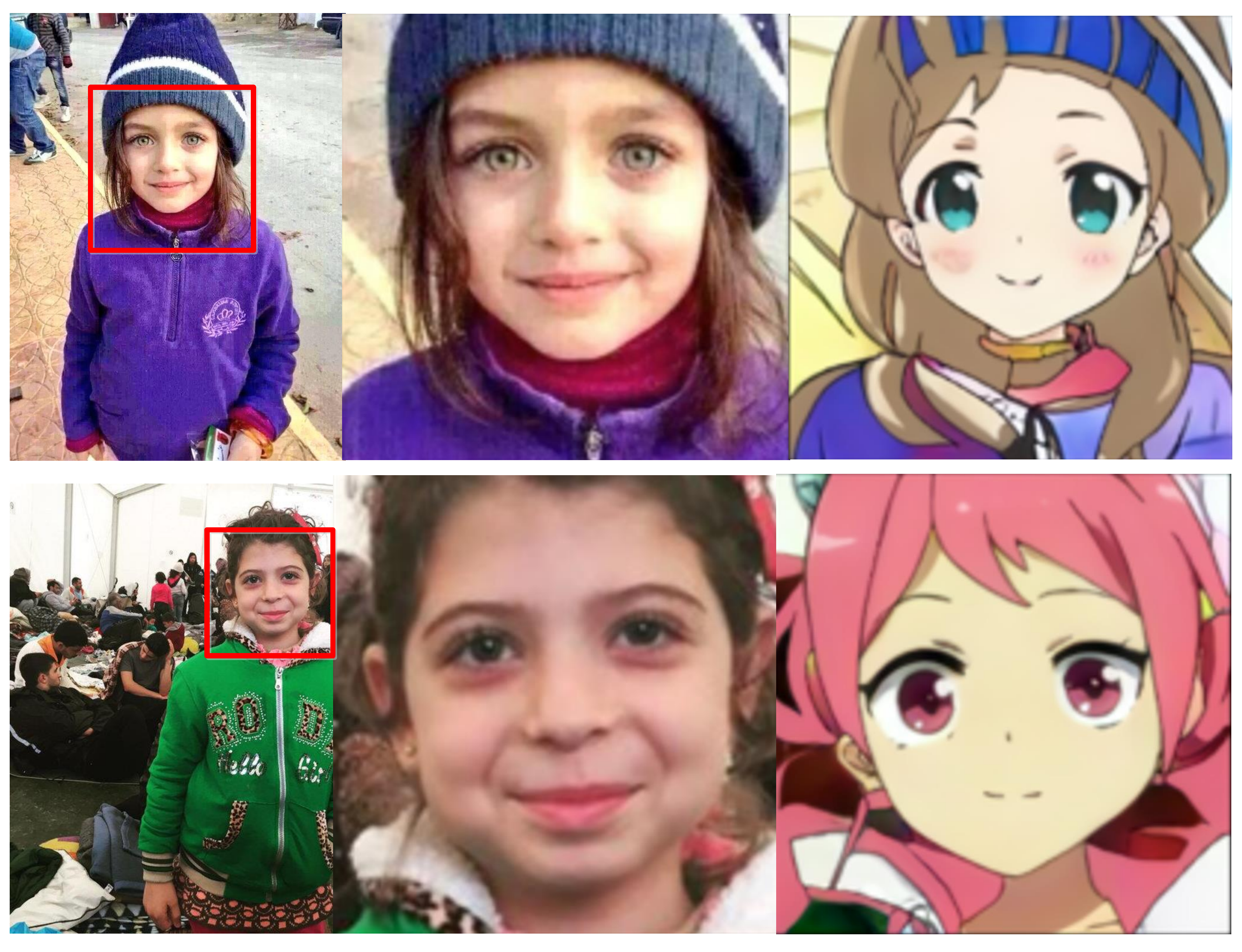}\\
  \caption{Style transferred examples.}
  \label{cartoon}
\end{figure}

\textbf{Implementation Details.}
We transform the images of people in fake news into cartoon style. CycleGAN first uses the CelebA face dataset and the first 50,000 random anime face datasets searched by google for 200 rounds of training. All images are converted to the size of 64$\times$64. The initial learning rates of the generator and discriminator are $10^{-4}$ and $4\times10^{-4}$ respectively. The images before and after the conversion are shown in Fig.~\ref{cartoon}. Then we feed these cartoon images into the detectors trained from clean examples for testing. We take the Twitter dataset and BDANN model as an example.

\textbf{Results and Analysis.}
We find that the accuracy of these cartoon images on clean detectors is surprisingly poor, reaching 36.30\%. It is concluded that the detectors will not work properly when tweets expressing the same meaning are converted into other image styles, which proves that the detectors are not robust enough in this respect.

\subsubsection{Case Where Images and Texts Do Not Correspond}\label{notcor}
\textbf{Implementation Details.}
In this section, we randomly scrambled the images corresponding to the tweets in the test set to express the inconsistency of the images and texts. The experiment process is similar to Section~\ref{image_style_transfer}. Firstly, the correspondence between images and texts is disrupted, and then put into the model trained from the clean examples. We experiment on Twitter data and BDANN model.

\begin{figure}[!t]
\centering
\subfloat[]{\includegraphics[width=1.1in]{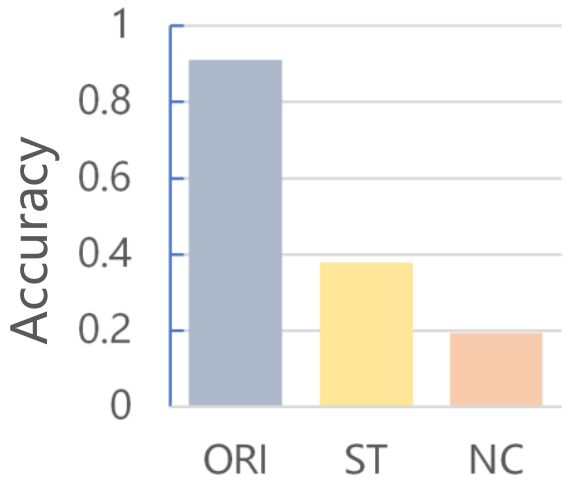}%
\label{case1}}
\hfil
\subfloat[]{\includegraphics[width=1.3in]{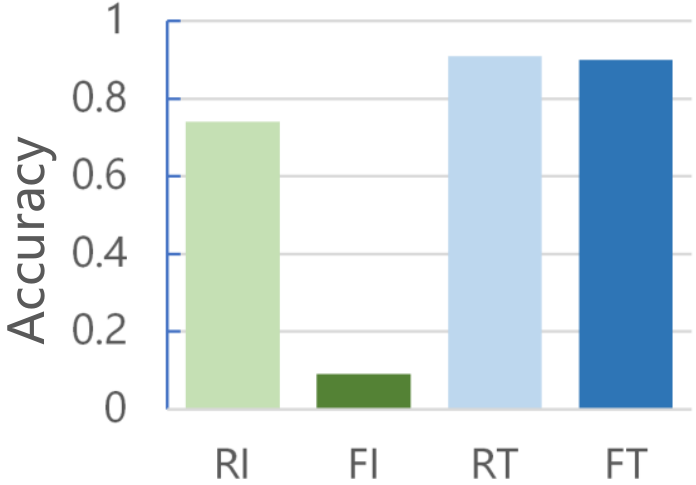}%
\label{case2}}
\caption{(a) The result of original images (ORI), image style transferred (ST) and the inconsistency of images and texts (NC). (b) The results of model bias evaluation, real images (RI), fake images (FI), real texts (RT), fake texts (FT).}
\label{image-style}
\end{figure}

\textbf{Results and Analysis.}
The results show that when the content is unchanged, the detectors cannot identify the tweets' authenticity where images and texts do not correspond. This suggests that we should not only pay attention to the performance improvement, but also to the connection between images and texts, such as semantic consistency. Fig.~\ref{image-style} (a) shows the performance when the images' style is transferred and the case where images and texts do not correspond.

% \begin{framed}
\textbf{Answer to RQ3:} (1) The visual modality of the multi-modal detectors is less robust, and the detection accuracy of the news containing the adversarial image with the same perturbation ratio drops more ($\epsilon=0.1$, the adversarial image drops by more than 30\%, the adversarial text drops less than 10\%); (2) The detector cannot correctly extract the features of the image after style transfer; (3) When the visual and textual information do not match, the detection performance of the detector decreases significantly.
% \end{framed}

\subsection{Robustness Evaluation of Model Bias}\label{bias}

In addition to adversarial attacks and backdoor attacks, we also conduct a bias evaluation on these detectors to evaluate whether the detectors rely on different features differently when making decisions. Inspired by~\cite{li2020shape} and the above results (the visual features cause greater damage to the detector), it is worth knowing whether the detectors are biased toward a specific feature, such as the visual feature.

When evaluating the text, we replace the real news with the texts of fake news and ensure that the image does not change. Meanwhile, we replace the fake news with the texts of real news. Then we use the models trained on the clean example to test it. The same process when evaluating the image. It's worth noting that when replacing, we do it in the same event, instead of randomly replacing other irrelevant content. 
% This is also the biggest difference from the experimental setup in Section~\ref{notcor}, which makes our synthetic tweets very confusing.

We test the fake category and the real category separately. The results are shown in Fig.~\ref{image-style} (b). Regarding the text, no matter what kind of replacement it is, it will not have much impact on the model. However, the replacement of images significantly impacts the model's performance, especially for the fake category. This means that the combination of fake text and the real image seems confusing to the detectors, reducing the accuracy to 6.44\%. This also shows that images seem to account for a large proportion of the detector's judgment of fake tweets. This further explains our conclusions in Section~\ref{adv} and Section~\ref{bkd}: compared with textual features, visual features are more susceptible to adversarial attacks and backdoor attacks, which greatly reduces the detectors' performance. This is because the detectors rely more on visual features, especially when making judgments on fake examples. 

% \begin{framed}
\textbf{Answer to RQ4:} The detection performance of multi-modal detectors can be improved using simple defense methods: (1) Image resize can improve the robustness of the detector against visual modality attacks imposed by malicious users (the accuracy can be improved by more than 30\%); (2) AC defense methods can improve detection robustness to visual modality attacks injected by malicious developers (the accuracy can reach more than 90\% of that in clean condition).
% \end{framed}

\section{Conclusion}\label{conclusion}
This work conducts a comprehensive evaluation of five multi-modal fake news detectors, including adversarial attacks, backdoor attacks, and bias evaluation. The results show that visual features are the common vulnerability of these detectors. We find the reason during the bias evaluation: the detectors rely more on visual features when making decisions, especially when judging fake news, which suggests researchers pay more attention to visual features when they improve the robustness of these detectors, especially the images corresponding to trending events. In addition, we find that the best-performing model is not necessarily the most robust. Considering the correlation between images and texts is also significantly important to improve the detectors' robustness. Finally, we defend against adversarial attacks and backdoor attacks on the visual features, respectively, which effectively improve the robustness of these detectors. 
% The experiment related data and code are available at
% \url{https://github.com/kenan976431/Robustness_Multi-modal_Detector}.

Our work is a preliminary exploration of these multi-modal fake news detectors' robustness. Several challenges remain, for example, we choose several classic attack and defense methods such as FGSM and image resizing to evaluate these detectors. In future works, we will try confrontation in more complex scenarios and more modal data (such as video and social context) to evaluate the detectors. In addition, fake news is often extremely provocative, leading its sentiment is often extreme. Therefore, we will also pay more attention to sentiment analysis in fake news detection tasks in future works, which may bring new possibilities to the robustness of these detectors.

% % use section* for acknowledgment
% \ifCLASSOPTIONcompsoc
%   % The Computer Society usually uses the plural form
%   \section*{Acknowledgments}
% \else
%   % regular IEEE prefers the singular form
% \fi

% The authors would like to thank the National Natural Science Foundation of China under Grant No. 62072406, the National Key Laboratory of Science and Technology on Information System Security under Grant No. 61421110502, the Key R\&D Projects in Zhejiang Province under Grant No. 2021C01117, the 2020 Industrial Internet Innovation Development Project under Grant No.TC200H01V and “Ten Thousand Talents Program” in Zhejiang Province under Grant No. 2020R52011.

% % Can use something like this to put references on a page
% % by themselves when using endfloat and the captionsoff option.
% \ifCLASSOPTIONcaptionsoff
%   \newpage
% \fi

\bibliography{ref}
\bibliographystyle{IEEEtran}

\end{document}